\pdfoutput=1

\documentclass[11pt,a4paper]{article}
\usepackage[hyperref]{acl2020}
\usepackage{times}
\usepackage{latexsym}

\usepackage{graphicx}
\usepackage{booktabs,multirow}
\usepackage{caption}
\usepackage[whole]{bxcjkjatype}
\usepackage{bbding}
\usepackage{pifont}
\usepackage{pifont}
\usepackage{subfig}
\usepackage{amssymb,mathtools}

\usepackage{bm}
\usepackage[normalem]{ulem} %
\def\dout{\bgroup
 \markoverwith{\lower-0.2ex\hbox
 {\kern-.03em\vbox{\hrule width.2em\kern0.45ex\hrule}\kern-.03em}}%
 \ULon}
\MakeRobust\dout
\usepackage{udline}
\usepackage{url}
\usepackage{xspace}
\usepackage{lingmacros}
\usepackage{tabularx}
\usepackage{longtable}
\usepackage{scrextend}

\aclfinalcopy %
\def\aclpaperid{648} %

\newcommand{\gl}[2]{%
\leavevmode\vtop{\hbox{#1}%
\hbox{#2\lower1.4ex\rlap{ }}}}

\usepackage{color}

\usepackage{microtype}

\title{Language Models as an Alternative Evaluator of Word Order Hypotheses: \\ A Case Study in Japanese}

\author{Tatsuki Kuribayashi$^{1,3}$, Takumi Ito$^{1,3}$, Jun Suzuki$^{1,2}$, Kentaro Inui$^{1,2}$  \\
 $^1$Tohoku University 
 $^2$RIKEN 
 $^3$Langsmith Inc. \\
 \texttt{\{kuribayashi, t-ito, jun.suzuki, inui\} @ecei.tohoku.ac.jp }}

\date{}

\begin{document}
\maketitle
\begin{abstract}
We examine a methodology using neural language models (LMs) for analyzing the word order of language.
This LM-based method has the potential to overcome the difficulties existing methods face, such as the propagation of preprocessor errors in count-based methods. 
In this study, we explore whether the LM-based method is valid for analyzing the word order.
As a case study, this study focuses on Japanese due to its complex and flexible word order. 
To validate the LM-based method, we test (i) parallels between LMs and human word order preference, and (ii) consistency of the results obtained using the LM-based method with previous linguistic studies. 
Through our experiments, we tentatively conclude that LMs display sufficient word order knowledge for usage as an analysis tool.
Finally, using the LM-based method, we demonstrate the relationship between the canonical word order and topicalization, which had yet to be analyzed by large-scale experiments.
\end{abstract}

\begin{table*}[t]
    \centering
    {\small
    \begin{tabular}{lcccccccc} \toprule
        & Topic & Time & Location & Subject & (Adverb) & Indirect object & Direct object & Verb \\ 
        \cmidrule(lr){1-1} \cmidrule(lr){2-2} \cmidrule(lr){3-3} \cmidrule(lr){4-4} \cmidrule(lr){5-5} \cmidrule(lr){6-6} \cmidrule(lr){7-7} \cmidrule(lr){8-8} \cmidrule(lr){9-9}
        Notation & \texttt{TOP} & \texttt{TIM} & \texttt{LOC} & \texttt{NOM} & - &  \texttt{DAT} & \texttt{ACC} & - \\ 
        Typical particle & ``{\it は}" ({\it wa}) & ``{\it に}" ({\it ni}) & ``{\it で}" ({\it de}) & ``{\it が}" ({\it ga}) & - & ``{\it に}" ({\it ni}) & ``{\it を}" ({\it o}) & - \\ %
        Related section & \ref{sec:topic} & \ref{subsec:place_time} & \ref{subsec:place_time} & \ref{subsec:place_time} & \ref{subsec:adverb} & \ref{subsec:cons_double} & \ref{subsec:cons_double} & \ref{subsec:cons_double}  \\ \bottomrule
        \end{tabular}
        }
        \caption{Overview of the typical cases in Japanese, their typical particles, and the sections where the corresponding case is analyzed. The well-known canonical word order of Japanese is listed from left to right.}
        \label{tbl:case_order}
\end{table*}

\section{Introduction}
\label{sec:intro}

Speakers sometimes have a range of options for word order in conveying a similar meaning.
A typical case in English is dative alternation:

\vspace{-0.15cm}
{\normalsize
\eenumsentence[1]{
    \item \shortexnt{1}{{\it A teacher gave a student a book.}}{}{}
    \vspace{-0.3cm}
    \item \shortexnt{1}{{\it A teacher gave a book to a student.}}{}{}
    }
}

\vspace{-0.15cm}
\noindent
Even for such a particular alternation, several studies~\cite{bresnan2007predicting, hovav2008english, colleman2009verb} investigated the factors determining this word order and found that the choice is not random.
For analyzing such linguistic phenomena, linguists repeat the cycle of constructing hypotheses and testing their validity, usually through psychological experiments or count-based methods.
However, these approaches sometimes face difficulties, such as scalability issues in psychological experiments and the propagation of preprocessor errors in count-based methods.

Compared to the typical approaches for evaluating linguistic hypotheses, approaches using LMs have potential advantages (Section~\ref{subsec:lm_diff}).
In this study, we examine the methodology of using LMs for analyzing word order (Figure~\ref{fig:intro}).
To validate the LM-based method, we first examine if there is a parallel between canonical word order and generation probability of LMs for each word order.
\citet{futrell2019rnns} reported that English LMs have human-like word order preferences, which can be one piece of evidence for validating the LM-based method.
However, it is not clear whether the above assumption is valid even in languages with more flexible word order.

{\setlength\textfloatsep{0pt}
\begin{figure}[t]
    \centering
      \includegraphics[width=7.5cm]{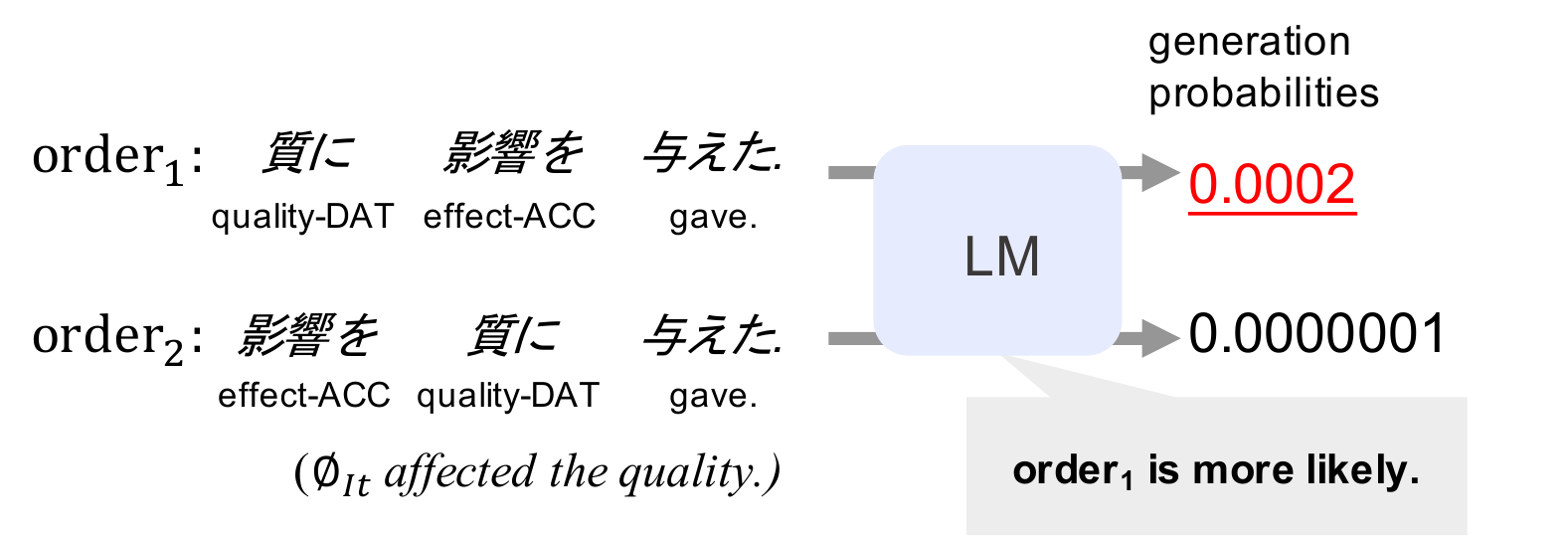}
      \vspace{-0.1cm}
      \caption{LM-based method for evaluating the canonicality of each word order considering their generation probabilities.} 
      \label{fig:intro}
\end{figure}
}

In this study, we specifically focus on the Japanese language due to its complex and flexible word order.
There are many claims on the canonical word order of Japanese, and it has attracted considerable attention from linguists and natural language processing (NLP) researchers for decades~\cite{hoji1985logical, saeki1998yosetsu, miyamoto2002sources, matsuoka2003two, koizumi2004cognitive, nakamoto2006, shigenaga2014canonical, sasano2016corpus, orita2017predicting, asahara-etal-2018-predicting}.

We investigated the validity of using Japanese LMs for canonical word order analysis by conducting two sets of experiments: (i) comparing word order preference in LMs to that in Japanese speakers (Section~\ref{sec:correl}), and (ii) checking the consistency between the preference of LMs with previous linguistic studies (Section~\ref{sec:consistency}).
From our experiments, we tentatively conclude that LMs display sufficient word order knowledge for usage as an analysis tool, and further explore potential applications.
Finally, we analyzed the relationship between topicalization and word order of Japanese by taking advantage of the LM-based method (Section~\ref{sec:topic}).

In summary, we:
\vspace{-0.2cm}
\begin{itemize}
\setlength{\parskip}{0cm} 
\setlength{\itemsep}{0.1cm}
\item Discuss and validate the use of LMs as a tool for word order analysis as well as investigate the sensitivity of LMs against different word orders in non-European language  (Section~\ref{sec:LM});
\item Find encouraging parallels between the results obtained with the LM-based method and those with the previously established method on various hypotheses of canonical word order of Japanese (Sections~\ref{sec:correl} and~\ref{sec:consistency}); and
\item Showcase the advantages of an LM-based method through analyzing linguistic phenomena that is difficult to explore with the previous data-driven methods (Section~\ref{sec:topic}).
\end{itemize}

\section{Linguistic background}
\label{sec:rel}

This section provides a brief overview of the linguistic background of canonical word order, some basics of Japanese grammar, and common methods of linguistic analysis.

\subsection{On canonical word order}
\label{subsec:rel_word_order}
Every language is assumed to have a canonical word order, even those with flexible word order~\cite{comrie1989language}.
There has been a significant linguistic effort to reveal the factors determining the canonical word order~\cite{bresnan2007predicting, hoji1985logical}.
The motivations for revealing the canonical word order range from linguistic interests to those involved in various other fields---it relates to language acquisition and production in psycholinguistics~\cite{slobin1982children, akhtar1999acquiring}, second language education~\cite{alonso2000teaching}, and natural language generation~\cite{visweswariah-etal-2011-word} or error correction~\cite{cheng-etal-2014-chinese} in NLP.
In Japanese, there are also many studies on its canonical word order~\cite{hoji1985logical, saeki1998yosetsu, koizumi2004cognitive, sasano2016corpus}.

\vspace{-0.1cm}
\paragraph*{Japanese canonical word order}
The word order of Japanese is basically subject-object-verb (SOV) order, but there is no strict rule except placing the verb at the end of the sentence~\cite{tsujimura2013introduction}.
For example, the following three sentences have the same denotational meaning ({\it ``A teacher gave a student a book."}):

\vspace{-0.15cm}
{\footnotesize
\eenumsentence[2]{
    \item \shortexnt{4}{{\it {\settensen \unc{先生が}}} & {\it \uwave{生徒に}}  & {\it \underline{本を}} & {\it {\bf あげた}}.} {teacher-\texttt{NOM} & student-\texttt{DAT} & book-\texttt{ACC} & gave.} {}
    \vspace{-0.3cm}
    \item \shortexnt{4}{{\it {\settensen \unc{先生が}}} & {\it \underline{本を}} & {\it \uwave{生徒に}} & {\it {\bf あげた}}.} {teacher-\texttt{NOM} &  book-\texttt{ACC} & student-\texttt{DAT} &  gave.} {}
    \vspace{-0.3cm}
    \item \shortexnt{4}{{\it \underline{本を}} & {\it \uwave{生徒に}}  & {\it {\settensen \unc{先生が}}} & {\it {\bf あげた}}.} {book-\texttt{ACC} & student-\texttt{DAT} & teacher-\texttt{NOM} & gave.} \\ {}
    }
}

\vspace{-0.3cm}
\noindent
This order-free nature suggests that the position of each constituent does not represent its semantic role (case). 
Instead, postpositional case particles indicate the roles. 
Table~\ref{tbl:case_order} shows typical constituents in a Japanese sentence, their postpositional particles, their canonical order, and the sections of this paper where each of them is analyzed.
Note that postpositional case particles are sometimes omitted or replaced with other particles such as adverbial particles (Section~\ref{sec:topic}).
These characteristics complicate the factors determining word order, which renders the automatic analysis of Japanese word order difficult.

\subsection{On typical methods for evaluating word order hypotheses and their difficulties}
\label{subsec:rel_eval_order}
There are two main methods in linguistic research: human-based methods, which observe human reactions, and data-driven methods, which analyze text corpora.

\vspace{-0.1cm}
\paragraph*{Human-based methods}
A typical approach of testing word order hypotheses is observing the reaction (e.g., reading time) of humans to each word order~\cite{shigenaga2014canonical, bahlmann2007fmri}.
These approaches are based on the direct observation of humans, but this method has scalability issues. %
There are also concerns that the participants may be biased, and that the experiments may not be replicable.

\vspace{-0.1cm}
\paragraph*{Data-driven methods}
Another typical approach is counting the occurrence frequencies of the targeted phenomena in a large corpus.
This count-based method is based on the assumption that there are parallels between the canonical word order and the frequency of each word order in a large corpus.
The parallel has been widely discussed~\cite{ARNON201067, bresnan2007predicting}, and many studies rely on this assumption~\cite{sasano2016corpus, kempen2004corpus}.
One of the advantages of this approach is suitability for large-scale experiments.
This enables considering a large number of examples.

In this method, researchers often have to identify the phenomena of interest with preprocessors (e.g., the predicate-argument structure parser used by~\citet{sasano2016corpus}) in order to count them.
However, sometimes, identification of the targeted phenomena is difficult for the preprocessors, which limits the possibilities of analysis.
For example, \citet{sasano2016corpus} focused only on simple examples where case markers appear explicitly, and only extract the head noun of the argument to avoid preprocessor errors.
Thus, they could not analyze the phenomena in which the above conditions were not met.
The above issue becomes more serious in low-resource languages, where the necessary preprocessors are often unavailable.

In this count-based direction, \citet{bloem2016testing} used n-gram LMs to test the claims on the German two-verb clusters.
This method is closest to our proposed approach, but the general validity of using LMs is out of focus.
This LM-based method also relies on the assumption of the parallels between the canonical word order and the frequency.

Another common data-driven approach is to train an interpretable model (e.g., Bayesian linear mixed models) to predict the targeted linguistic phenomena and analyze the inner workings of the model (e.g., slope parameters)~\cite{bresnan2007predicting, asahara-etal-2018-predicting}.
Through this approach, researchers can obtain richer statistics, such as the strength of each factor's effect on the targeted phenomena, but creating labeled data and designing features for supervised learning can be costly.

\section{LM-based method}
\label{sec:LM}

\subsection{Overview of the LM-based method}
In the NLP field, LMs are widely used to estimate the acceptability of text~\cite{olteanu2006language, kann2018sentence}.
An overview of the LM-based method is shown in Figure~\ref{fig:intro}.
After preparing several word orders considering the targeted linguistic hypothesis, we compare their generation probabilities in LMs.
We assume that the word order with the highest generation probability follows their canonical word order.

\subsection{Advantages of the LM-based method}
\label{subsec:lm_diff}

In the count-based methods mentioned in Section~\ref{subsec:rel_eval_order}, researchers often require preprocessors to identify the occurrence of the phenomena of interest in a large corpus.
On the other hand, researchers need to prepare data to be scored by LMs to evaluate hypothesis in the LM-based method.
Whether it is easier to prepare the preprocessor or the evaluation data depends on the situation.
For example, the data preparation is easier in the situation where one wants to analyze the word order trends when a specific postpositional particle is omitted.
The question is whether Japanese speakers prefer the word order like in Example (3)-a or (3)-b.\footnote{Omitted characters are crossed out. (e.g., \dout{を})}

\vspace{-0.1cm}
{\footnotesize
\eenumsentence[3]{
    \item \shortexnt{3}{{\it {\settensen \unc{生徒に}}} & {\it \underline{本\dout{を}}} & {\it {\bf あげた}}.} {student-\texttt{DAT} & book(-\texttt{ACC}) & gave.} {}
    \vspace{-0.3cm}
    \item \shortexnt{3}{{\it \underline{本\dout{を}}} & {\it {\settensen \unc{生徒に}}} & {\it {\bf あげた}}.} {book(-\texttt{ACC}) & student-\texttt{DAT} & gave.} {}
}
}

\vspace{-0.2cm}
\noindent
While identifying the cases (\texttt{ACC} in Example (3)) without their postpositional particle is difficult, creating the data without a specific postpositional particle by modifying the existing data is easier such as creating Example (4)-b from Example (4)-a.

\vspace{-0.1cm}
{\footnotesize
\eenumsentence[4]{
    \item \shortexnt{3}{{\it {\settensen \unc{生徒に}}} & {\it \underline{本を}} & {\it {\bf あげた}}.} {student-\texttt{DAT} & book-\texttt{ACC} & gave.} {}
    \vspace{-0.3cm}
    \item \shortexnt{3}{{\it {\settensen \unc{生徒に}}} & {\it \underline{本\dout{を}}} & {\it {\bf あげた}}.} {student-\texttt{DAT} & book(-\texttt{ACC}) & gave.} {}
}
}

\vspace{-0.3cm}
\noindent
Thus, in such situation, the LM-based method can be suitable.

The human-based method is more reliable given an example.
However, it can be prohibitively costly.
While the human-based method requires an evaluation data and human subjects, the LM-based method only requires the evaluation data. 
Thus, the LM-based method can be more suitable for estimating the validity of hypotheses and considering many examples as exhaustively as possible.
In addition, the LM-based method can be replicable.
The suitable approach can be different in a situation, and broadening the choice of alternative methodologies may be beneficial to linguistic research.

Nowadays, various useful frameworks, language resources, and machine resources required to train LMs are available,\footnote{For example, one can train LMs with fairseq~\cite{ott2019fairseq} and Wikipedia data on cloud computing platforms.} which support the ease of implementing the LM-based method.
Moreover, we make the LMs used in this study available.\footnote{\url{https://github.com/kuribayashi4/LM_as_Word_Order_Evaluator}.}

\subsection{Strategies to validate the use of LM to analyze the word order}
\label{subsec:lm_valid}
The goal of this study is to validate the use of LMs for analyzing the canonical word order.
The canonical word order itself is still a subject of research, and the community does not know all about it.
Thus, it is ultimately impossible to enumerate the requirements on what LMs should know about the canonical word order and probe the knowledge of LMs. 
Instead, we demonstrate the validity of the LM-based method by showcasing two types of parallels: (i) word order preference of LMs showing parallels with that of humans, and (ii) the results obtained with the LM-based method and those with previous methods being consistent on various claims on canonical word order.
If the results of LMs are consistent with those of existing methods, the possibility that LMs and existing methods have the same ability to evaluate the hypotheses is supported. 
If the LM-based method is assumed to be valid, the method has the potential to streamline the research on unevaluated claims on word order.
In the experiment sections, we examine the properties of Japanese LMs on (i) and (ii).

\subsection{CAUTION -- when using LMs for evaluating linguistic hypotheses}
\label{subsec:lm_causion}
Even if LMs satisfy the criteria described in~\ref{subsec:lm_valid}, there is no exact guarantee that LM scores will reflect the effectiveness of human processing of specific constructions in general.
Thus, there seems to be a danger of confusing LM artifacts with language facts.
Based on this, we hope that researchers use LMs as a tool just to limit the hypothesis space.
LM supported hypotheses should then be re-verified with a human-based approach. 

Furthermore, since there is a lot of hypotheses and corresponding research, we cannot check all the properties of LMs in this study.
This study focuses on intra-sentential factors of Japanese case order, and it is still unclear whether the LM-based method works properly in linguistic phenomena which are far from being the focus of this study. 
This is the first study where evidence is collected on the validity of using LMs for word order analysis and encourages further research on collecting such evidence and examining under what conditions this validity is guaranteed.

\subsection{LMs settings}
\label{subsec:lm_p}
We used auto-regressive, unidirectional LMs with Transformer~\cite{vaswani2017}.
We used two variants of LMs, a character-based LM (CLM) and a subword-based LM (SLM).
In training SLM, the input sentences are once divided into morphemes by MeCab~\cite{kudo2006mecab} with a UniDic dictionary,\footnote{\url{https://unidic.ninjal.ac.jp/}} and then these morphemes are split into subword units by byte-pair-encoding.~\cite{sennrich-etal-2016-neural}\footnote{Implemented in sentencepiece~\cite{kudo2018sentencepiece} We set character coverage to 0.9995，and vocab size to 100,000.}.
160M sentences\footnote{14GB in UTF-8 encoding. For reference, Japanese Wikipedia has around 2.5 GB of text. Because the focus of this study has context-independent nature, the sentences order is shuffled to prevent learning the inter-sentential characteristics of the language.} randomly selected from 3B web pages were used to train the LMs.
Hyperparameters are shown in Appendix~\ref{sec:appendix_hyper}.

Given a sentence $s$, we calculate its generation probability $p(s)=\overrightarrow{p}(s)\cdot \overleftarrow{p}(s)$,
where $\overrightarrow{p}(\cdot)$ and $\overleftarrow{p}(\cdot)$ are generation probabilities calculated by a left-to-right LM and a right-to-left LM, respectively.
Depending on the hypothesis, we compare the generation probabilities of various variants of $s$ with different word orders.
We assume that the word order with the highest generation probability follows their canonical word order.

\section{Experiment1: comparing human and LMs word order preference}
\label{sec:correl}

{\setlength\textfloatsep{0pt}
\begin{figure}[t]
    \centering
      \includegraphics[width=7.5cm]{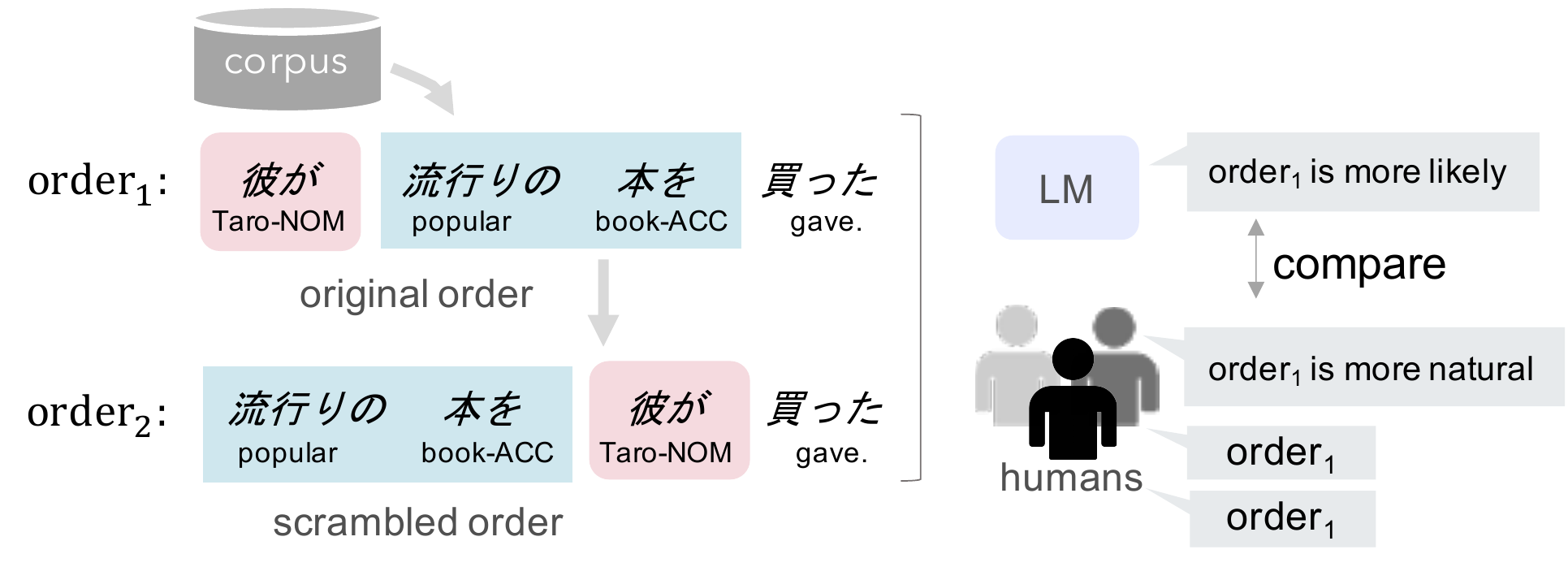}
      \vspace{-0.1cm}
      \caption{Overview of the experiment of comparing human and LMs word order preference. First, we created data for the task of comparing the appropriateness of the word order (left part), then we compare the preference of LMs and humans through this task (right part).}
      \label{fig:assess}
\end{figure}
}

To examine the validity of using LMs for canonical word order analysis, we examined the parallels between the LMs and humans on the task determining the canonicality of the word order (Figure~\ref{fig:assess}).
First, we created data for this task (Section \ref{subsec:correl_data}).
We then compared the word order preference of LMs and that of humans (Section~\ref{subsec:correl_result}).

\noindent
\subsection{Human annotation}
\label{subsec:correl_data}

\paragraph*{Data}
We randomly collected 10k sentences from 3B web pages, which are not overlapped with the LM training data.
To remove overly complex sentences, we extracted sentences that must: (i) have less than or equal to five clauses and one verb, (ii) have clauses with a sibling relationship in its dependency tree, and they accompany a particle or adverb, (iii) not have special symbols such as parentheses, and (iv) not have a backward dependency path.
For each sentence, we created its scrambled version.\footnote{When several scrambled versions were possible for a given sentence, we randomly selected one of them.}
The scrambling process is as follows:

\vspace {-0.3 cm}
\begin {enumerate}
\setlength { \parskip} {0 cm}
\setlength { \itemsep} {0 cm}
\item Identify the dependency structure by using JUMAN\footnote{\url{http://nlp.ist.i.kyoto-u.ac.jp/EN/index.php?JUMAN}} and KNP\footnote{\url{http://nlp.ist.i.kyoto-u.ac.jp/EN/index.php?KNP}}. %
\item Randomly select a clause with several children.
\item Shuffle the position of its children along with their descendants.
\end {enumerate}

\vspace {-0.3 cm}
\noindent

\vspace{-0.2cm}
\paragraph*{Annotation}
We used the crowdsourcing platform Yahoo Japan!\footnote{\url{https://crowdsourcing.yahoo.co.jp/}}. %
For our task, we showed crowdworkers a pair of sentences (order$_1$, order$_2$), where one sentence has the original word order, and the other sentence has a scrambled word order.\footnote{Crowdworkers did not know which sentence was the original sentence.}
Each annotator was instructed to label the pair with one of the following choices: (1) order$_1$ is better, (2) order$_2$ is better, or (3) the pair contains a semantically broken sentence. 
Only the sentences (order$_1$, order$_2$) were shown to the annotators, and they were instructed not to imagine a specific context for the sentences.
We filtered unmotivated workers by using check questions.\footnote{We manually created check questions considering the Japanese speakers' preference in trial experiments in advance.}
For each pair instance, we employed 10 crowdworkers. 
In total, 756 unique, motivated crowdworkers participated in our task.

From the annotated data, we collected only the pairs satisfying the following conditions for our experiments: (i) none of 10 annotators determined that the pair contains a semantically broken sentence, and (ii) nine or more annotators preferred the same order.
The majority decision is labeled in each pair; the task is binary classification.
We assume that if many workers prefer a certain word order, then it follows its canonical word order, and the other one deviates from it.
We collected 2.6k pair instances of sentences.

\subsection{Result}
\label{subsec:correl_result}

We compared the word order preference of LMs and that of the workers by using the 2.6K pairs created in Section~\ref{subsec:correl_data}.
We calculated the correlation of the decisions between the LMs and the workers; which word order is more appropriate order$_1$ or order$_2$.
The word orders supported by CLM and SLM are highly correlated with workers, with the Pearson correlation coefficient of 0.89 and 0.90, respectively.
This supports the assumption that the generation probability of LMs can determine the canonical word order as accurately as humans do.
Note that such a direct comparison of word order is difficult with the count-based methods because of the sparsity of the corpus.

\section{Experiment2: consistency with previous studies}
\label{sec:consistency}
This section examines whether LMs show word order preference consistent with previous linguistic studies.
The results are entirely consistent, which support the validity of the LM-based methods in Japanese.
Each subsection focuses on a specific component of Japanese sentences.

\subsection{Double objects}
\label{subsec:cons_double}

The order of double objects is one of the most controversial topics in Japanese word order. %
Examples of the possible order are as follows:

{\small
\begin{description}
\setlength{\itemindent}{-15.5pt}
\item[\textmd{(5)}] \begin{description}
    \item[\textmd{\texttt{DAT-ACC}:}] \gl{{\it \underline{生徒に}}}{student-\texttt{DAT}} \gl{{\it \underline{本を}}}{book-\texttt{ACC}} \gl{{\it {\bf あげた}}}{gave.}
    \item[\textmd{\texttt{ACC-DAT}:}] \gl{{\it \underline{本を}}}{book-\texttt{ACC}} \gl{{\it \uwave{生徒に}}}{student-\texttt{DAT}} \gl{{\it {\bf あげた}}}{gave.}
\end{description}
\end{description}
}

\vspace{-0.2cm}
\noindent
Henceforth, \texttt{DAT-ACC}$\,$/$\,$\texttt{ACC-DAT} denotes the word order in which the  \texttt{DAT}$\,$/$\,$\texttt{ACC} argument precedes the \texttt{ACC}$\,$/$\,$\texttt{DAT} argument.
We evaluate the claims~\citet{sasano2016corpus} focused on with the data they collected.\footnote{We filtered the examples overlapping with the training data of LMs in advance. As a result, we collected 4.5M examples.}

{\setlength\textfloatsep{0pt}
\begin{figure}[t]
    \centering
    \subfloat[][Each verb's \texttt{ACC-DAT} rate.]{
        \includegraphics[width=3.5cm]{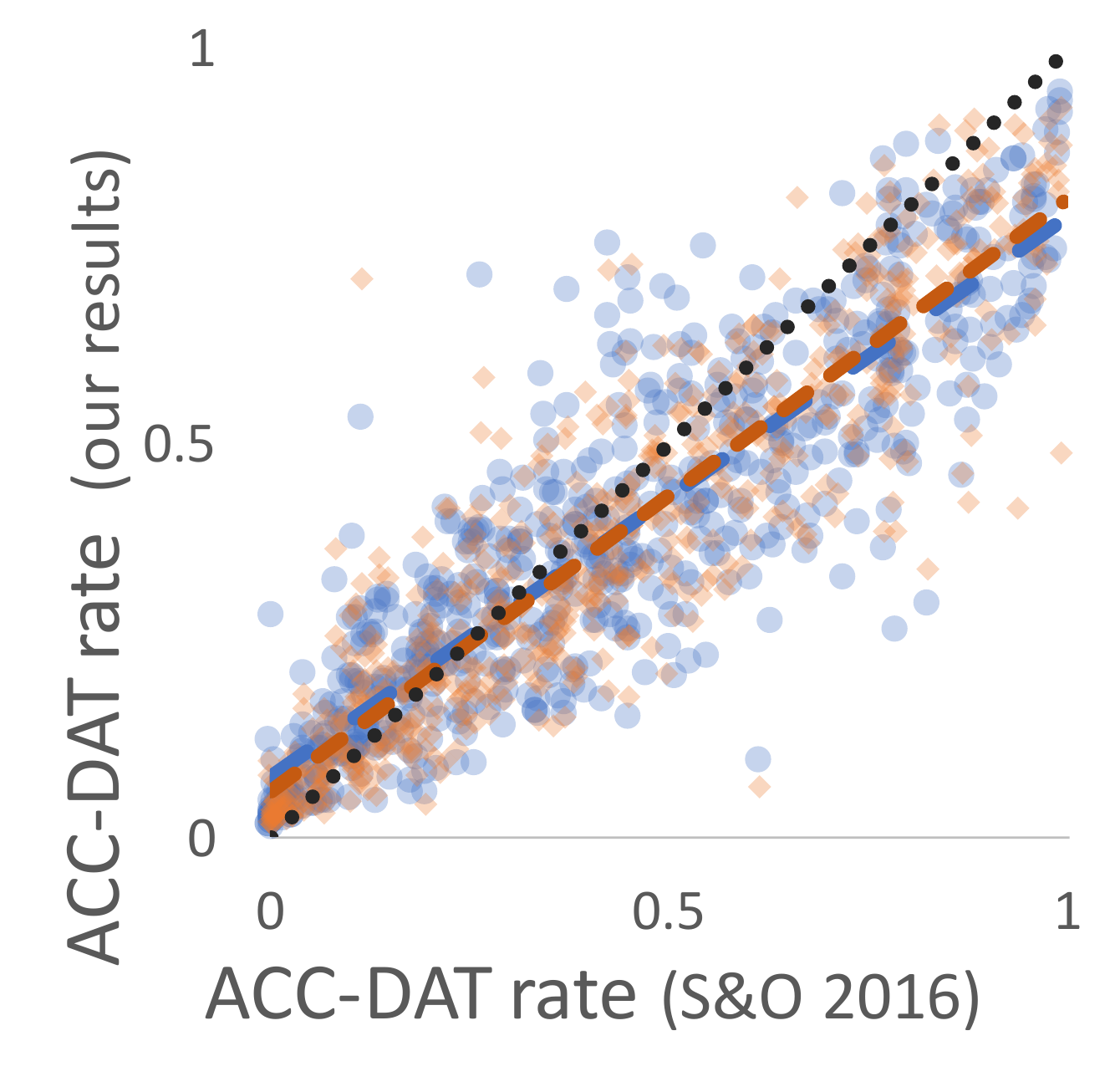}
        \label{fig:verb620}
    }
    \hfill
    \subfloat[][Relationship between  \\each verb's $R^v_{\text{\texttt{DAT}-only}}$ and \\ the \texttt{ACC-DAT} rate.]{
        \includegraphics[width=3.5cm]{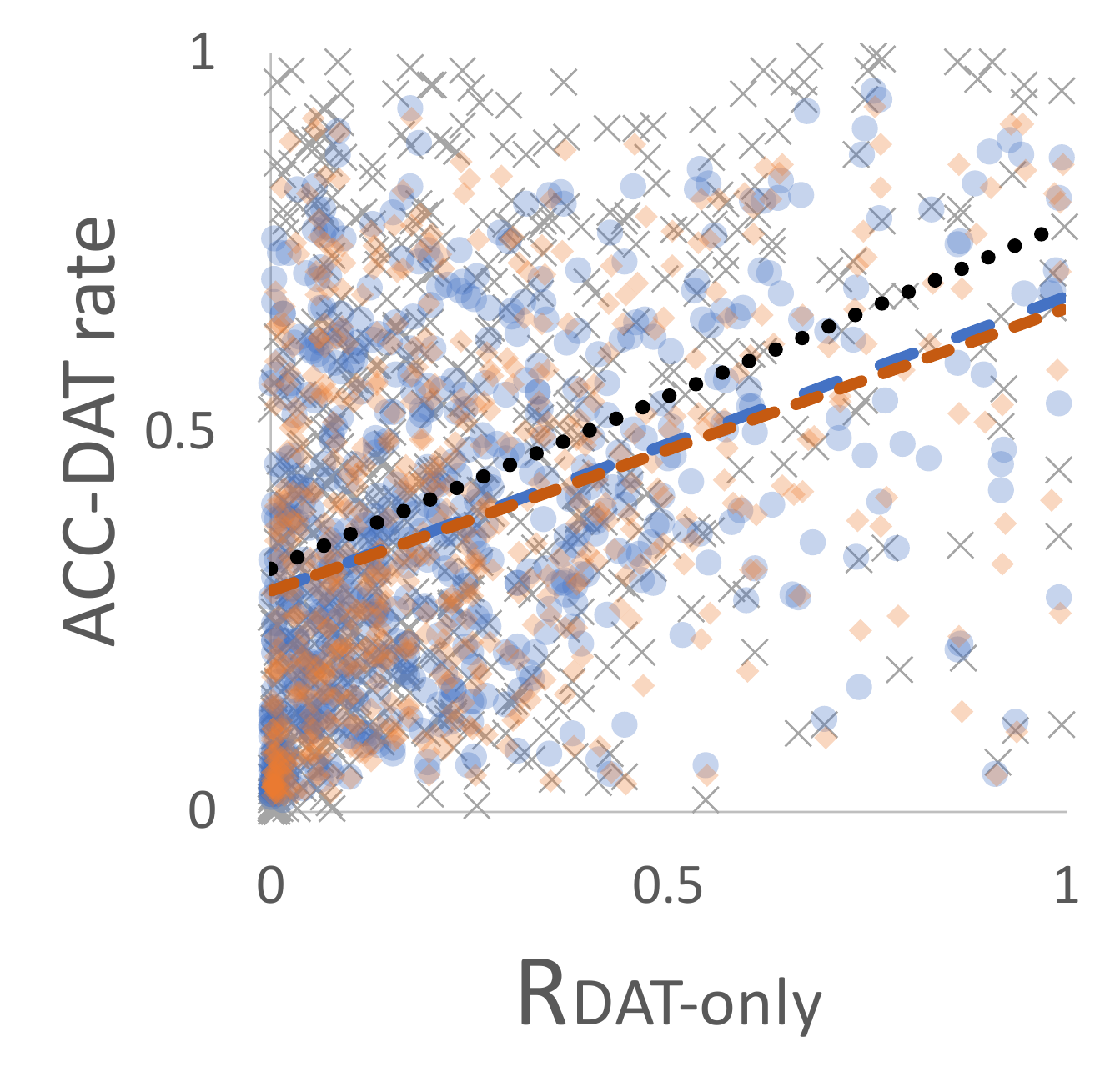}
        \label{fig:omission}
    }
    \\
    \subfloat[][Relationship between the degree of co-occurrence of verb and arguments, and the \texttt{ACC-DAT} rate in each example. For the results of LMs, the \texttt{ACC-DAT} rate of each example is regarded as 1 if LMs prefer \texttt{ACC-DAT} order, otherwise we regard the example as 0.]{
        \includegraphics[width=7.5cm]{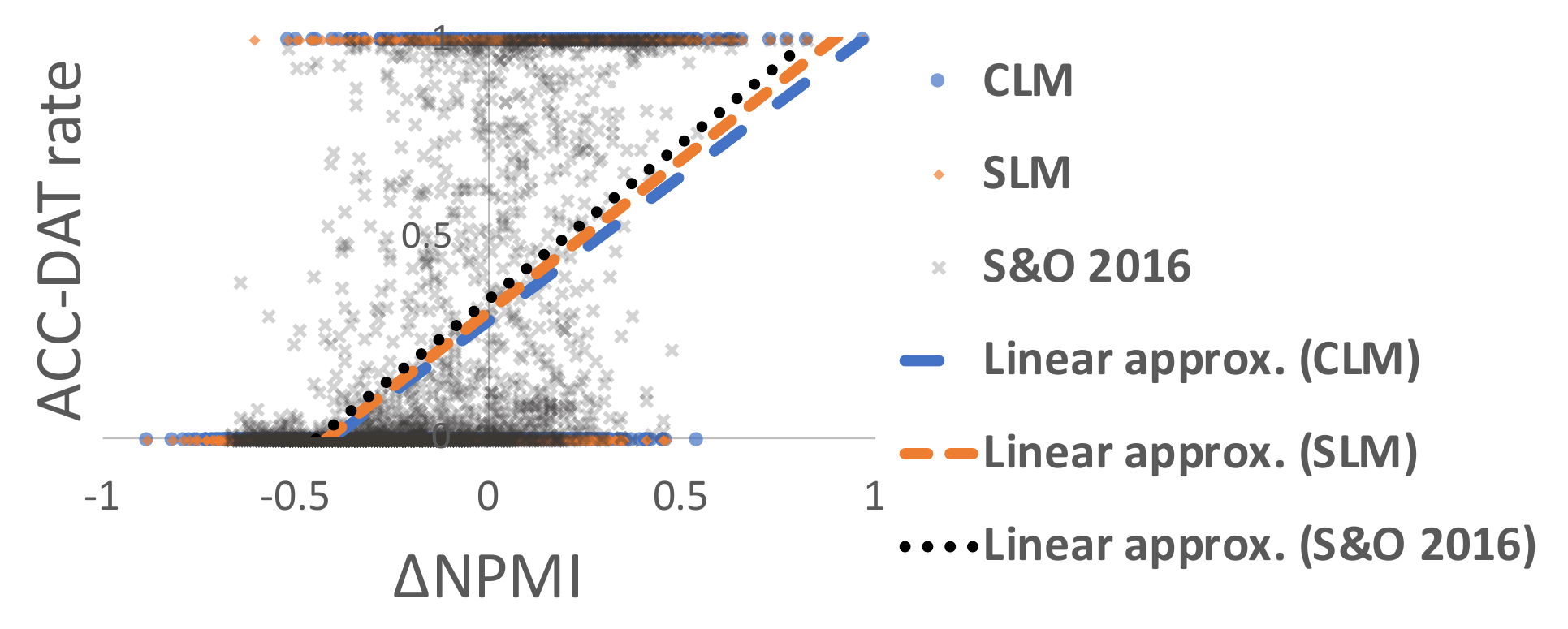}
        \label{fig:co-occur}
    }
      \caption{Overlap of the results of~\citet{sasano2016corpus} and that of LMs. In figures (a) and (b), each plot corresponds to each verb. In figure (c), each plot corresponds to each example.　The legend of figure (a) and (b) is the same as in figure (c). ``S\&O 2016'' refers to~\citet{sasano2016corpus}.%
      } 
     \label{fig:double_obj}
\end{figure}
}

\vspace{-0.1cm}
\paragraph*{Word order for each verb}
\label{subsubsec:cons_verb}
\noindent
First, we analyzed the trend of the double object order for each verb.
We analyzed 620 verbs following~\citet{sasano2016corpus}.\footnote{We removed verbs for which all examples overlap with the data for training the LMs.}
For each set of examples $S^v$ corresponding to a verb $v$, we: (i) created an instance with the swapped order of \texttt{ACC} and \texttt{DAT} for each example, and (ii) compared the generation probabilities of the original and swapped instance.
$\hat{S}^v$ is the set of examples preferred by LMs.
$R_{\texttt{ACC-DAT}}^v$ is calculated as follows:

\vspace{-0.4cm}
\begin{align*}
\small
R_{\texttt{ACC-DAT}}^v &= \frac{N^v_{\texttt{ACC-DAT}}}{N^v_{\texttt{ACC-DAT}} + N^v_{\texttt{DAT-ACC}}} \:\:,
\end{align*}

\noindent
where $N^v_{\texttt{ACC-DAT}}\,/\,N^v_{\texttt{DAT-ACC}}$ is the number of examples with the \texttt{ACC-DAT}$\,$/$\,$\texttt{DAT-ACC} order in $\hat{S}^v$. 

Figure~\ref{fig:double_obj}-(a) shows the relationship between $R_{\texttt{ACC-DAT}}^v$ determined by LMs and one reported in a previous count-based study~\cite{sasano2016corpus}.
These results strongly correlate with the Pearson correlation coefficient of 0.91 and 0.88, in CLM and SLM, respectively.
In addition, ``canonical word order is \texttt{DAT-ACC}"~\cite{hoji1985logical} is unlikely to be valid because there are verbs where $R_{\texttt{ACC-DAT}}^v$ is very high (details in Appendix~\ref{subsec:appendix_word_order_and_each_verb}).
This conclusion is consistent with~\citet{sasano2016corpus}. 

\vspace{-0.1cm}
\paragraph*{Word order and verb types}
\label{subsubsec:cons_verb_type}
\noindent
In Japanese, there are show-type and pass-type verbs (details in Appendix~\ref{subsec:appendix_word_order_and_verb_type}).
\citet{matsuoka2003two} claimed that the order of double objects differs depending on these verb types.
Following~\citet{sasano2016corpus}, we analyzed this trends.

We applied the Wilcoxon rank-sum test between the distributions of $R_{\texttt{ACC-DAT}}^v$ determined by LMs in the two groups (show-type  and pass-type verbs).
The results show no significant difference between the two groups (p-value is 0.17 and 0.12 in the experiments using CLM and SLM, respectively).
These results are consistent with the count-based~\cite{sasano2016corpus} and the human-based~\cite{miyamoto2002sources, koizumi2004cognitive} methods.

\vspace{-0.1cm}
\paragraph*{Word order and argument omission}
\label{subsubsec:cons_ommision}
\citet{sasano2016corpus} claimed that the frequently omitted case is placed near the verb.
First, we calculated $R^v_{\text{\texttt{DAT}-only}}$ for each verb $v$ as follows:

\vspace{-0.4cm}
\begin{align*} 
\small
R_{\texttt{DAT}\text{-only}}^{v} &= \frac{N^v_{\texttt{DAT}\text{-only}}}{N^v_{\texttt{DAT}\text{-only}} + N^v_{\texttt{ACC}\text{-only}}} \:\:,
\end{align*}

\vspace{-0.2cm}
\noindent
where $N^v_{\texttt{DAT}\text{-only}}\,/\,N^v_{\texttt{ACC}\text{-only}}$ denotes the number of examples in which the \texttt{DAT}$\,$/$\,$\texttt{ACC} case appears, and the other case does not in $S^v$.
A large  $R^v_{\text{\texttt{DAT}-only}}$ score indicates that the \texttt{DAT} argument is less frequently omitted than the \texttt{ACC} argument in $S^v$.
We analyzed the relationship between $R^v_{\text{\texttt{DAT}-only}}$ and $R_{\texttt{ACC-DAT}}^v$ for each verb.

Figure~\ref{fig:double_obj}-(b) shows that the regression lines from the LM-based method and~\citet{sasano2016corpus} corroborate similar trends.
The Pearson correlation coefficient between $R^v_{\text{\texttt{DAT}-only}}$ and $R_{\texttt{ACC-DAT}}^v$ is 0.404 for CLM and 0.374 for SLM.
The results are consistent with ~\citet{sasano2016corpus}, where they  reported that the correlation coefficient was 0.391.

\vspace{-0.1cm}
\paragraph*{Word order and semantic role of the dative argument}
\label{subsubsec:cons_animacy}
 \citet{matsuoka2003two} claimed that the canonical word order differs depending on the semantic role of the dative argument.
 \citet{sasano2016corpus} evaluated this claim by analyzing the trend in the following two types of examples:

{\small
\begin{description}
\setlength{\itemindent}{-15.5pt}
\item[\textmd{(6)}] \begin{description}
    \item[\textmd{Type-A:}] \gl{{\it \underline{本を}}}{book-\texttt{ACC}} \gl{{\it \uwave{学校に}}}{school-\texttt{DAT}} \gl{{\it {\bf 返した}}}{returned.}
    \item[\textmd{Type-B:}] \gl{{\it \uwave{先生に}}}{teacher-\texttt{DAT}} \gl{{\it \underline{本を}}}{book-\texttt{ACC}} \gl{{\it {\bf 返した}}}{returned.}
\end{description}
\end{description}
}

\vspace{-0.3cm}
\noindent
Type-A has an inanimate goal ({\it school}) as the \texttt{DAT} argument, while Type-B has an animate processor ({\it teacher}).
It was reported that Type-A is likely to be the \texttt{ACC-DAT} order, while Type-B is likely to be the \texttt{DAT-ACC} order.
Following~\citet{sasano2016corpus}, we analyzed 113 verbs.\footnote{Among the 126 verbs used in~\citet{sasano2016corpus}, 113 verbs with data that do not overlap with the LM training data were selected.}
For each verb, we compared the \texttt{ACC-DAT} rate in its type-A examples and the rate in its type-B examples.

The number of verbs where the \texttt{ACC-DAT} order is preferred in Type-A examples to Type-B examples is significantly larger (a two-sided sign test p $<$ 0.05).
This result is consistent with that of \citet{sasano2016corpus, matsuoka2003two} and implies that the LMs capture the animacy of the nouns. 
Details are in Appendix~\ref{subsec:appendix_word_order_and_semantic_role_dative_argument}.

\vspace{-0.1cm}
\paragraph*{Word order and co-occurrence of verb and arguments}
\citet{sasano2016corpus} claimed that an argument that frequently co-occurs with the verb tends to be placed near the verb.
For each example, the LMs determine which word order (\texttt{DAT-ACC} or \texttt{ACC-DAT}) is appropriate.
Each example also has a score $\Delta$NPMI (definition in Appendix~\ref{subsec:appendix_word_order_and_co-occurrence_of_verb_and_arguments}).
Higher $\Delta$NPMI means that the \texttt{DAT} noun in the example more strongly co-occurs with the verb in the example than the \texttt{ACC} noun.

Figure~\ref{fig:double_obj}-(c) shows the relationship between $\Delta$NPMI and the \texttt{ACC-DAT} rate in each example.
$\Delta$NPMI and the \texttt{ACC-DAT} rate are correlated with the Pearson correlation coefficient of 0.517 and 0.521 in CLM and SLM, respectively.
These results are consistent with~\citet{sasano2016corpus}.

\begin{table}[t]
    \centering
    \renewcommand{\arraystretch}{0.9}
    {\small
    \begin{tabular}{lccc} \toprule
        & \texttt{TIM}$<$\texttt{LOC} &  \texttt{TIM}$<$\texttt{NOM} & \texttt{LOC}$<$\texttt{NOM} \\ 
        \cmidrule(lr){1-1} \cmidrule(lr){2-2} \cmidrule(lr){3-3} \cmidrule(lr){4-4}
        CLM & .757 & .642 & .604 \\ 
        SLM & .708 & .632 & .615  \\ 
        Count & .686 & .666 & .681  \\ \bottomrule
        \end{tabular}
        }
        \caption{The columns $a<b$ show the score $o(a<b)$, which indicates the rate of case $a$ being more likely to be placed before $b$. The row ``Count'' shows the count-based results in the dataset we used.}
        \label{tbl:order_time_place}
\end{table}

\subsection{Order of constituents representing time, location, and subject information}
\label{subsec:place_time}

Our focus moves to the cases closer to the beginning of the sentences.
The following claim is a well-known property of Japanese word order: ``The case representing time information (\texttt{TIM}) is placed before the case representing location information (\texttt{LOC}), and the \texttt{TIM} and \texttt{LOC} cases are placed before the \texttt{NOM} case"~\cite{saeki1960, saeki1998yosetsu}.
We examined a parallel between the result obtained with the LM-based and count-based methods on this claim.

We randomly collected 81k examples from 3B web pages.\footnote{Without overlap with the training data of LMs.}
To create the examples, we identified the case components by KNP, and the \texttt{TIM} and \texttt{LOC} cases were categorized with JUMAN (details in Appendix~\ref{sec:appendix_Wikipedia_data}).
For each example $s$, we created all possible word orders and obtained the word order with the highest generation probability ($\hat{s}$).
Given $\hat{S}$ a set of $\hat{s}$, we calculated a score $o (a < b)$ for cases $a$ and $b$ as follows:

\vspace{-0.5cm}
\begin{align*}
\small
o(a < b) & = \frac{N_{a<b}}{N_{a<b} + N_{b<a}} \:\:,
\end{align*}

\vspace{-0.2cm}
\noindent
where $N_{k<l}$ is the number of examples where the case $k$ precedes the case $l$ in $\hat{S}$.
Higher $o(a<b)$ indicates that the case {\it a} is more likely to be placed before the case {\it b}.
The results with the LM-based methods and the count-based method are consistent (Table~\ref{tbl:order_time_place}).
Both results show that $o(\texttt{TIM}\:<\:\texttt{LOC})$ is significantly larger than $o(\texttt{TIM}\:>\:\texttt{LOC})$  ($p<0.05$ with a two-sided signed test), which indicates that the \texttt{TIM} case usually precedes the \texttt{LOC} case.
Similarly, the results indicate that the \texttt{TIM} case and the \texttt{LOC} case precedes the \texttt{NOM} case.

\subsection{Adverb position}
\label{subsec:adverb}

\begin{table}[t]
    \centering
    \renewcommand{\arraystretch}{0.8}
    {\small
    \begin{tabular}{lcccc} \toprule
         Model & \textsc{Modal} & \textsc{Time} & \textsc{Manner} & \textsc{Resultive} \\ 
        \cmidrule(lr){1-1} \cmidrule(lr){2-2} \cmidrule(lr){3-3} \cmidrule(lr){4-4} \cmidrule(lr){5-5}
        CLM  & 1. & 1 & 0.5 & 1. \\
        SLM & 1. & 0.5 & 1. & 0.5 \\ \bottomrule
        \end{tabular}
        }
        \caption{The scores denote the rank correlation between the preference of each adverb position in LMs and that reported in~\cite{koizumi2006}.}
        \label{tbl:adverb}
\end{table}

We checked the preference of the adverb position in LMs.
The position of the adverb has no restriction except that it must be before the verb, which is similar to the trend of the case position.
However, \citet{koizumi2006} claimed that ``There is a canonical position of an adverb depending on its type."
They focus on four types of adverbs: \textsc{modal}, \textsc{time}, \textsc{manner}, and \textsc{resultive}.

We used the same examples as~\citet{koizumi2006}.
For each example $s$, we created its three variants with a different adverb position as follows (``{\it A friend handled the tools \underline{roughly}.}"):

{\small
\begin{description}
\item[\textmd{(10)}] \begin{description}
    \item[ASOV: ] \gl{{\it \underline{乱暴に}}}{roughly} \gl{{\it 友達が}}{friend-\texttt{NOM}} \gl{{\it 道具を}}{tools-\texttt{ACC}}  \gl{{\it {\bf 扱った}}}{handled.}
    \item[SAOV: ] \gl{{\it 友達が}}{friend-\texttt{NOM}} \gl{{\it \underline{乱暴に}}}{roughly} \gl{{\it 道具を}}{tools-\texttt{ACC}}  \gl{{\it {\bf 扱った}}}{handled.}
    \item[SOAV: ] \gl{{\it 友達が}}{friend-\texttt{NOM}} \gl{{\it 道具を}}{tools-\texttt{ACC}} \gl{{\it \underline{乱暴に}}}{roughly}  \gl{{\it {\bf 扱った}}}{handled.}
\end{description}
\end{description}
}

\vspace{-0.15cm}
\noindent
where the sequence of the alphabet such as ``ASOV'' denote the word order of its corresponding sentences. 
For example, ``ASOV'' indicates the order: adverb $<$ subject $<$ object $<$ verb.
``A,'' ``S,'' ``O,'' and ``V'' denote ``adverb,'' ``subject,'' ``object,'' and ``verb,'' respectively. 

Then, we obtained the preferred adverb position by comparing their generation probabilities.
Finally, for each adverb type and its examples, we ranked the preference of the possible adverb positions: ``ASOV,'' ``SAOV,'' and ``SOAV.'' %
Table~\ref{tbl:adverb} shows the rank correlation of the preference of the position of each adverb type.
The results show similar trends of LMs with that of the human-based method~\cite{koizumi2006}.

\subsection{Long-before-short effect}
\label{subsec:length}

\begin{table}[t]
    \centering
    \renewcommand{\arraystretch}{0.8}
    {\small
    \begin{tabular}{lccc} \toprule
        Model & long precedes short & short precedes long \\ 
        \cmidrule(lr){1-1} \cmidrule(lr){2-2} \cmidrule(lr){3-3}
        CLM & 5,640 & 3,754 \\ 
        SLM & 5,757 & 3,914  \\  \bottomrule
        \end{tabular}
        }
        \caption{Changes in the position of a constituent with the largest number of chunks. }
        \label{tbl:length}
\end{table}

The effects of ``long-before-short,'' the trend that a long constituent precedes a short one, has been reported in several studies~\cite{asahara-etal-2018-predicting, orita2017predicting}．
We checked whether this effect can be captured with the LM-based method.
Among the examples used in Section~\ref{subsec:place_time}, we analyzed about 9.5k examples in which the position of the constituent with the largest number of chunks\footnote{chunks were identified by KNP.} differed between its canonical case order\footnote{In this section, canonical case order is assumed to be \texttt{TOM}$<$\texttt{LOC}$<$\texttt{NOM}$<$\texttt{DAT}$<$\texttt{ACC}.} and the order supported by LMs.

Table~\ref{tbl:length} shows that there are significantly (p $<$ 0.05 with a two-sided signed test) large numbers of examples where the longest constituent moves closer to the beginning of the sentence. 
This result is consistent with existing studies and supports the tendency for longer constituents to appear before shorter ones.

\subsection{Summary of the results}
\label{subsec:summary}
We found parallels between the results with the LM-based method and that with the previously established method on various properties of canonical word order.
These results support the use of LMs for analyzing Japanese canonical word order.

\section{Analysis: word order and topicalization}
\label{sec:topic}

In the previous section, we tentatively concluded that LMs can be used for analyzing the intra-sentential properties on the canonical word order.
Based on this finding, in this section, we demonstrate the analysis of additional claims on the properties of the canonical word order with the LM-based method, which has been less explored by large-scale experiments.
This section shows the analysis of the relationship between topicalization and the canonical word order.
Additional analyses on the effect of various adverbial particles for the word order are shown in Appendix~\ref{sec:appendix_add_ana}.

\subsection{Topicalization in Japanese}

The adverbial particle ``{\it は}" (\texttt{TOP}) is usually used as a postpositional particle when a specific constituent represents the topic or focus of the sentence~\cite{heycock1993syntactic, noda1996, fry2003ellipsis}.
When a case component is topicalized, the constituent moves to the beginning of the sentence, and the particle ``{\it は}" (\texttt{TOP}) is added~\cite{noda1996}.
Additionally, the original case particle is sometimes omitted,\footnote{The particles ``{\it を}" (\texttt{ACC}) and ``{\it が}" (\texttt{NOM}) are omitted.} which makes the case of the constituent difficult to identify.
For example, to topicalize ``{\it 本を}" ({\it book}-\texttt{ACC}) in Example (8)-a, the constituent moves to the beginning of the sentence, and the original accusative case particle ``{\it を}" (\texttt{ACC}) is omitted. Similarly, ``{\it 先生が}" ({\it teacher}-\texttt{NOM}) is topicalized in Example (8)-b. 
The original sentence is enclosed in the square brackets in Example (8). 

\vspace{-0.15cm}
{\small
\eenumsentence[8]{
    \item \shortexnt{4}{{\it \underline{本\dout{を}は}} & [{\it 先生が} & \dout{{\it \underline{本を}}} & {\it {\bf あげた}}.]} {book-\texttt{TOP} & teacher-\texttt{NOM} & \dout{book-\texttt{ACC}} & gave.}  \\{}
    \vspace{-0.3cm}
    \item \shortexnt{4}{{\it \underline{先生\dout{が}は}} & [\dout{{\it 先生が}} & {\it \underline{本を}} & {\it {\bf あげた}}.]} {teacher-\texttt{TOP} & \dout{teacher-\texttt{NOM}} & book-\texttt{ACC} & gave.}  \\{}
}
}

\vspace{-0.3cm}
\noindent
With the above process, we can easily create a sentence with a topicalized constituent.
On the other hand, identifying the original case of the topicalized case components is error-prone.
Thus, the LM-based method can be suitable for empirically evaluating the claims related to the topicalization.

\subsection{Experiments and results}
\label{subsec:topic:exp}
By using the LM-based method, we evaluate the following two claims: 

\vspace {-0.2 cm}
\begin{description}
\setlength { \parskip} {0 cm}
\setlength { \itemsep} {0 cm}
\item[(i)] The more anterior the case is in the canonical word order, the more likely its component is topicalized~\cite{noda1996}.
\item[(ii)] The more the verb prefers the \texttt{ACC-DAT} order, the more likely the \texttt{ACC} case is topicalized than the \texttt{DAT} case.
\end{description}

\vspace {-0.2 cm}
\noindent
The claim (i) suggests that, for example, the \texttt{NOM} case is more likely to be topicalized than the \texttt{ACC} case because the \texttt{NOM} case is before the \texttt{ACC} case in the canonical word order of Japanese.
The claim (ii) is based on our observation.
It can be regarded as an extension of the claim (i) considering the effect of the verb on its argument order.
We assume that the canonical word order of Japanese is \texttt{TIM}$\:<\:$\texttt{LOC}$\:<\:$\texttt{NOM}$\:<\:$\texttt{DAT}$\:<\:$\texttt{ACC} in this section.

\vspace{-0.1cm}
\paragraph{Claim (i)}
We examine which case is more likely to be topicalized.
We collected 81k examples from Japanese Wikipedia (Details are in Appendix~\ref{sec:appendix_Wikipedia_data}).
For each example, a set of candidates was created by topicalizing each case, as shown in Example (8). 
Then, we selected the sentences with the highest score by LMs in each candidate set.
We denote the obtained sentences as $\hat{S}^{\text{topic}}$.
We calculated a score $t_{a|b}$ for pairs of cases $a$ and $b$.

{\normalsize
\vspace{-0.5cm}
\begin{align*}
t_{a|b} & = \frac{N_{a|b}}{N_{a|b} + N_{b|a}}
\end{align*}
}

\vspace{-0.4cm}
\noindent
where $N_{a|b}$ is the examples where the case $a$ and $b$ appear, and case $a$ is a topic of the sentence in $\hat{S}^{\text{topic}}$.
The higher the score is, the more the case $a$ is likely to be topicalized than the case $b$ is.

We compared  $t_{a|b}$ and $t_{b|a}$ among the pairs of cases $a$ and $b$, where the case $a$ precedes the case $b$ in the canonical word order.
Through our experiments, $t_{a|b}$ was significantly larger than $t_{b|a}$ ($p<0.05$ with a paired t-test) in CLM and SLM results, which supports the claim (i)~\cite{noda1996}. 
Detailed results are shown in Appendix~\ref{sec:appendix_detail_ana}.

\vspace{-0.1cm}
\paragraph{Claim (ii)}
The canonical word order of double objects is different for each verb  (Section~\ref{subsec:cons_double}).
Based on this assumption and the claim (i), we hypothesized that the more the verb prefers the \texttt{ACC-DAT} order, the more likely the \texttt{ACC} case of the verb is topicalized than the \texttt{DAT} case.

We used the same data as in Section~\ref{subsec:cons_double}.
For each example, we created two sentences by topicalizing the \texttt{ACC} or \texttt{DAT} argument.
Then we compared their generation probabilities.
In each set of examples corresponding to a verb $v$, we calculated the rate that the sentence with the topicalized \texttt{ACC} argument is preferred rather than that with the topicalized \texttt{DAT} argument.
This rate and $R_{\texttt{ACC-DAT}}^v$ is significantly correlated with the  Pearson correlation coefficient of 0.89 and 0.84 in CLM and SLM, respectively.
This results support the claim (ii). 
Detailed results are shown in Appendix~\ref{sec:appendix_detail_ana}.

\section{Conclusion and Future work}
\label{sec:conclusion}
We have proposed to use LMs as a tool for analyzing word order in Japanese. 
Our experimental results support the validity of using Japanese LMs for canonical word order analysis, which has the potential to broaden the possibilities of linguistic research.
From an engineering view, this study supports the use of LMs for scoring Japanese word order automatically.
From the viewpoint of the linguistic field, we provide additional empirical evidence to various word order hypotheses as well as demonstrate the validity of the LM-based method.

We plan to further explore the capability of LMs on other linguistic phenomena related to word order, such as ``given new ordering''~\cite{nakagawa2016information,asahara-etal-2018-predicting}.
Since LMs are language-agnostic, analyzing word order in another language with the LM-based method would also be an interesting direction to investigate.
Furthermore, we would like to extend a comparison between machine and human language processing beyond the perspective of word order. 

\section{Acknowledgments}
We would like to offer our gratitude to Kaori Uchiyama for taking the time to discuss our paper and Ana Brassard for her sharp feedback on English. 
We also would like to show our appreciation to the Tohoku NLP lab members for their valuable advice. We are particularly grateful to Ryohei Sasano for sharing the data for double objects order analyses. 
This work was supported by JST CREST Grant Number JPMJCR1513, JSPS KAKENHI Grant Number JP19H04162, and Grant-in-Aid for JSPS Fellows Grant Number JP20J22697.

\newpage

\bibliography{acl2020}

\begin{thebibliography}{39}
\expandafter\ifx\csname natexlab\endcsname\relax\def\natexlab#1{#1}\fi

\bibitem[{Akhtar(1999)}]{akhtar1999acquiring}
Nameera Akhtar. 1999.
\newblock \href {https://doi.org/10.1017/S030500099900375X} {{Acquiring basic
  word order: Evidence for data-driven learning of syntactic structure}}.
\newblock \emph{Journal of child language}, 26(2):339--356.

\bibitem[{Alonso~Belmonte et~al.(2000)}]{alonso2000teaching}
Isabel Alonso~Belmonte et~al. 2000.
\newblock \href {http://hdl.handle.net/10017/952} {{Teaching English Word Order
  to ESL Spanish Students. A Functional Perspective}}.

\bibitem[{Arnon and Snider(2010)}]{ARNON201067}
Inbal Arnon and Neal Snider. 2010.
\newblock \href {https://doi.org/https://doi.org/10.1016/j.jml.2009.09.005}
  {{More than words: Frequency effects for multi-word phrases}}.
\newblock \emph{Journal of Memory and Language}, 62(1):67--82.

\bibitem[{Asahara et~al.(2018)Asahara, Nambu, and
  Sano}]{asahara-etal-2018-predicting}
Masayuki Asahara, Satoshi Nambu, and Shin-Ichiro Sano. 2018.
\newblock \href {https://doi.org/10.18653/v1/W18-2805} {{Predicting Japanese
  Word Order in Double Object Constructions}}.
\newblock In \emph{Proceedings of the Eight Workshop on Cognitive Aspects of
  Computational Language Learning and Processing}, pages 36--40, Melbourne.
  Association for Computational Linguistics.

\bibitem[{Bahlmann et~al.(2007)Bahlmann, Rodriguez-Fornells, Rotte, and
  M{\"u}nte}]{bahlmann2007fmri}
J{\"o}rg Bahlmann, Antoni Rodriguez-Fornells, Michael Rotte, and Thomas~F
  M{\"u}nte. 2007.
\newblock \href {https://doi.org/10.1002/hbm.20318} {{An fMRI study of
  canonical and noncanonical word order in German}}.
\newblock \emph{Human brain mapping}, 28(10):940--949.

\bibitem[{Bloem(2016)}]{bloem2016testing}
Jelke Bloem. 2016.
\newblock \href {https://www.aclweb.org/anthology/W16-4120} {{Testing the
  Processing Hypothesis of word order variation using a probabilistic language
  model}}.
\newblock In \emph{Proceedings of the Workshop on Computational Linguistics for
  Linguistic Complexity (CL4LC)}, pages 174--185, Osaka, Japan. The COLING 2016
  Organizing Committee.

\bibitem[{Bresnan et~al.(2007)Bresnan, Cueni, Nikitina, and
  Baayen}]{bresnan2007predicting}
Joan Bresnan, Anna Cueni, Tatiana Nikitina, and R~Harald Baayen. 2007.
\newblock \href {https://pure.mpg.de/pubman/item/item_58830_2} {{Predicting the
  dative alternation}}.
\newblock In \emph{Cognitive foundations of interpretation}, pages 69--94.
  KNAW.

\bibitem[{Cheng et~al.(2014)Cheng, Yu, and Chen}]{cheng-etal-2014-chinese}
Shuk-Man Cheng, Chi-Hsin Yu, and Hsin-Hsi Chen. 2014.
\newblock \href {https://www.aclweb.org/anthology/C14-1028} {{Chinese Word
  Ordering Errors Detection and Correction for Non-Native Chinese Language
  Learners}}.
\newblock In \emph{Proceedings of COLING 2014, the 25th International
  Conference on Computational Linguistics: Technical Papers}, pages 279--289,
  Dublin, Ireland. Dublin City University and Association for Computational
  Linguistics.

\bibitem[{Colleman(2009)}]{colleman2009verb}
Timothy Colleman. 2009.
\newblock \href {https://doi.org/https://doi.org/10.1016/j.langsci.2008.01.001}
  {{Verb disposition in argument structure alternations: a corpus study of the
  dative alternation in Dutch}}.
\newblock \emph{Language Sciences}, 31(5):593--611.

\bibitem[{Comrie(1989)}]{comrie1989language}
Bernard Comrie. 1989.
\newblock \href
  {https://www.press.uchicago.edu/ucp/books/book/chicago/L/bo24426144.html}
  {\emph{{Language universals and linguistic typology: Syntax and
  morphology}}}.
\newblock University of Chicago press.

\bibitem[{Fry(2003)}]{fry2003ellipsis}
John Fry. 2003.
\newblock \href {https://doi.org/10.4324/9780203484036} {\emph{{Ellipsis and
  wa-marking in Japanese conversation}}}.
\newblock Taylor \& Francis.

\bibitem[{Futrell and Levy(2019)}]{futrell2019rnns}
Richard Futrell and Roger~P Levy. 2019.
\newblock \href {https://doi.org/10.7275/jb34-9986} {{Do RNNs learn human-like
  abstract word order preferences?}}
\newblock In \emph{Proceedings of the Society for Computation in Linguistics
  (SCiL) 2019}, pages 50--59.

\bibitem[{Grave et~al.(2017)Grave, Joulin, Ciss{\'e}, Grangier, and
  J{\'e}gou}]{armand-adaptive-softmax}
{\'E}douard Grave, Armand Joulin, Moustapha Ciss{\'e}, David Grangier, and
  Herv{\'e} J{\'e}gou. 2017.
\newblock \href {http://proceedings.mlr.press/v70/grave17a.html} {{Efficient
  softmax approximation for GPUs}}.
\newblock In \emph{Proceedings of the 34th International Conference on Machine
  Learning}, volume~70 of \emph{Proceedings of Machine Learning Research},
  pages 1302--1310, International Convention Centre, Sydney, Australia. PMLR.

\bibitem[{Heycock(1993)}]{heycock1993syntactic}
Caroline Heycock. 1993.
\newblock \href {https://doi.org/10.1007/BF01732503} {{Syntactic predication in
  Japanese}}.
\newblock \emph{Journal of East Asian Linguistics}, 2(2):167--211.

\bibitem[{Hoji(1985)}]{hoji1985logical}
Hajime Hoji. 1985.
\newblock \href {https://books.google.co.jp/books?id=HgTdnQEACAAJ} {{Logical
  form constraints and configurational structures in Japanese.}}
\newblock \emph{PHD Thesis. University of Washington}.

\bibitem[{Hovav and Levin(2008)}]{hovav2008english}
Malka~Rappaport Hovav and Beth Levin. 2008.
\newblock \href {https://doi.org/10.1017/S0022226707004975} {{The English
  dative alternation: The case for verb sensitivity}}.
\newblock \emph{Journal of linguistics}, 44(1):129--167.

\bibitem[{Kann et~al.(2018)Kann, Rothe, and Filippova}]{kann2018sentence}
Katharina Kann, Sascha Rothe, and Katja Filippova. 2018.
\newblock \href {https://doi.org/10.18653/v1/K18-1031} {{Sentence-Level Fluency
  Evaluation: References Help, But Can Be Spared!}}
\newblock In \emph{Proceedings of the 22nd Conference on Computational Natural
  Language Learning}, pages 313--323, Brussels, Belgium. Association for
  Computational Linguistics.

\bibitem[{Kempen and Harbusch(2004)}]{kempen2004corpus}
Gerard Kempen and Karin Harbusch. 2004.
\newblock \href {https://doi.org/10.1515/9783110894028.173} {{A corpus study
  into word order variation in German subordinate clauses: Animacy affects}}.
\newblock \emph{Multidisciplinary approaches to language production}, pages
  173--181.

\bibitem[{Koizumi and Tamaoka(2004)}]{koizumi2004cognitive}
Masatoshi Koizumi and Katsuo Tamaoka. 2004.
\newblock \href {https://doi.org/10.11435/gengo1939.2004.125_173} {{Cognitive
  processing of Japanese sentences with ditransitive verbs}}.
\newblock \emph{Gengo Kenkyu (Journal of the Linguistic Society of Japan)},
  2004(125):173--190.

\bibitem[{Koizumi and Tamaoka(2006)}]{koizumi2006}
Masatoshi Koizumi and Katsuo Tamaoka. 2006.
\newblock \href {https://doi.org/10.11225/jcss.13.392} {{The Canonical
  Positions of Adjuncts in the Processing of Japanese Sentence}}.
\newblock \emph{Cognitive Studies: Bulletin of the Japanese Cognitive Science
  Society}, 13(3):392--403.

\bibitem[{Kudo(2006)}]{kudo2006mecab}
Taku Kudo. 2006.
\newblock {Mecab: Yet another part-of-speech and morphological analyzer}.
\newblock \emph{http://mecab.sourceforge.jp}.

\bibitem[{Kudo and Richardson(2018)}]{kudo2018sentencepiece}
Taku Kudo and John Richardson. 2018.
\newblock \href {https://doi.org/10.18653/v1/D18-2012} {{SentencePiece: A
  simple and language independent subword tokenizer and detokenizer for Neural
  Text Processing}}.
\newblock In \emph{Proceedings of the 2018 Conference on Empirical Methods in
  Natural Language Processing: System Demonstrations}, pages 66--71, Brussels,
  Belgium. Association for Computational Linguistics.

\bibitem[{Matsuoka(2003)}]{matsuoka2003two}
Mikinari Matsuoka. 2003.
\newblock \href {https://doi.org/10.1023/A:1022472109327} {{Two Types of
  Ditransitive Consturctions in Japanese}}.
\newblock \emph{Journal of East Asian Linguistics}, 12(2):171--203.

\bibitem[{Miyamoto(2002)}]{miyamoto2002sources}
Edson~T Miyamoto. 2002.
\newblock \href {https://ci.nii.ac.jp/naid/10015080761/en/} {{Sources of
  difficulty in the processing of scrambling in Japanese}}.
\newblock \emph{Sentence processing in East Asian languages}, pages 167--188.

\bibitem[{Nakagawa(2016)}]{nakagawa2016information}
Natsuko Nakagawa. 2016.
\newblock Information structure in spoken japanese: Particles, word order, and
  intonation.

\bibitem[{Nakamoto et~al.(2006)Nakamoto, Lee, and Kuroda}]{nakamoto2006}
Keiko Nakamoto, Jae-ho Lee, and Kow Kuroda. 2006.
\newblock \href {https://doi.org/10.11225/jcss.13.334} {{Preferred Word Orders
  Correlate with “Sentential” Meanings That Cannot Be Reduced to Verb
  Meanings: A New Perspective on “Construction Effects” in Japanese}}.
\newblock \emph{Cognitive Studies: Bulletin of the Japanese Cognitive Science
  Society}, 13(3):334--352.

\bibitem[{Noda(1996)}]{noda1996}
Hisashi Noda. 1996.
\newblock \href {http://www.9640.jp/book_view/?128} {\emph{{Wa to ga [Wa and
  ga]}}}.
\newblock Kurosio Publishers.

\bibitem[{Olteanu et~al.(2006)Olteanu, Suriyentrakorn, and
  Moldovan}]{olteanu2006language}
Marian Olteanu, Pasin Suriyentrakorn, and Dan Moldovan. 2006.
\newblock \href {https://www.aclweb.org/anthology/W06-3122} {Language models
  and reranking for machine translation}.
\newblock In \emph{Proceedings of the Workshop on Statistical Machine
  Translation}, pages 150--153, New York City. Association for Computational
  Linguistics.

\bibitem[{Orita(2017)}]{orita2017predicting}
Naho Orita. 2017.
\newblock \href {https://www.aclweb.org/anthology/W17-0706/} {Predicting
  japanese scrambling in the wild}.
\newblock In \emph{Proceedings of the 7th Workshop on Cognitive Modeling and
  Computational Linguistics}, pages 41--45.

\bibitem[{Ott et~al.(2019)Ott, Edunov, Baevski, Fan, Gross, Ng, Grangier, and
  Auli}]{ott2019fairseq}
Myle Ott, Sergey Edunov, Alexei Baevski, Angela Fan, Sam Gross, Nathan Ng,
  David Grangier, and Michael Auli. 2019.
\newblock \href {https://doi.org/10.18653/v1/N19-4009} {{fairseq: A Fast,
  Extensible Toolkit for Sequence Modeling}}.
\newblock In \emph{Proceedings of the 2019 Conference of the North American
  Chapter of the Association for Computational Linguistics (Demonstrations)},
  pages 48--53, Minneapolis, Minnesota. Association for Computational
  Linguistics.

\bibitem[{Saeki(1960)}]{saeki1960}
Tetsuo Saeki. 1960.
\newblock \href {https://ci.nii.ac.jp/naid/40001065096/} {{Gendaigo ni okeru
  gojun no keik{\=o} -- iwayuru hogo no baai [The trend of word order in modern
  writing-- in so-called complements]}}.
\newblock \emph{Gengo seikatsu [Language life]}, (111):56--63.

\bibitem[{Saeki(1998)}]{saeki1998yosetsu}
Tetsuo Saeki. 1998.
\newblock \href {http://www.9640.jp/book_view/?165} {\emph{{Y{\=o}setsu Nihongo
  no Gojun [Essentials of Japanese word order]}}}.
\newblock Kurosio Publishers.

\bibitem[{Sasano and Okumura(2016)}]{sasano2016corpus}
Ryohei Sasano and Manabu Okumura. 2016.
\newblock \href {https://doi.org/10.18653/v1/P16-1211} {{A Corpus-Based
  Analysis of Canonical Word Order of Japanese Double Object Constructions}}.
\newblock In \emph{Proceedings of the 54th Annual Meeting of the Association
  for Computational Linguistics (Volume 1: Long Papers)}, pages 2236--2244,
  Berlin, Germany. Association for Computational Linguistics.

\bibitem[{Sennrich et~al.(2016)Sennrich, Haddow, and
  Birch}]{sennrich-etal-2016-neural}
Rico Sennrich, Barry Haddow, and Alexandra Birch. 2016.
\newblock \href {https://doi.org/10.18653/v1/P16-1162} {{Neural Machine
  Translation of Rare Words with Subword Units}}.
\newblock In \emph{Proceedings of the 54th Annual Meeting of the Association
  for Computational Linguistics (Volume 1: Long Papers)}, pages 1715--1725,
  Berlin, Germany. Association for Computational Linguistics.

\bibitem[{Shigenaga(2014)}]{shigenaga2014canonical}
Yasumasa Shigenaga. 2014.
\newblock \href {https://doi.org/10.7575/aiac.alls.v.5n.2p.35} {{Canonical Word
  Order of Japanese Ditransitive Sentences: A Preliminary Investigation through
  a Grammaticality Judgment Survey}}.
\newblock \emph{Advances in Language and Literary Studies}, 5(2):35--45.

\bibitem[{Slobin and Bever(1982)}]{slobin1982children}
Dan~I Slobin and Thomas~G Bever. 1982.
\newblock \href {https://doi.org/10.1016/0010-0277(82)90033-6} {{Children use
  canonical sentence schemas: A crosslinguistic study of word order and
  inflections}}.
\newblock \emph{Cognition}, 12(3):229--265.

\bibitem[{Tsujimura(2013)}]{tsujimura2013introduction}
Natsuko Tsujimura. 2013.
\newblock \href
  {https://www.wiley.com/en-us/An+Introduction+to+Japanese+Linguistics%2C+3rd+Edition-p-9781444337730}
  {\emph{{An introduction to Japanese linguistics}}}.
\newblock John Wiley \& Sons.

\bibitem[{Vaswani et~al.(2017)Vaswani, Shazeer, Parmar, Uszkoreit, Jones,
  Gomez, Kaiser, and Polosukhin}]{vaswani2017}
Ashish Vaswani, Noam Shazeer, Niki Parmar, Jakob Uszkoreit, Llion Jones,
  Aidan~N Gomez, \L~ukasz Kaiser, and Illia Polosukhin. 2017.
\newblock \href
  {http://papers.nips.cc/paper/7181-attention-is-all-you-need.pdf} {{Attention
  is All you Need}}.
\newblock In I.~Guyon, U.~V. Luxburg, S.~Bengio, H.~Wallach, R.~Fergus,
  S.~Vishwanathan, and R.~Garnett, editors, \emph{Advances in Neural
  Information Processing Systems 30}, pages 5998--6008. Curran Associates, Inc.

\bibitem[{Visweswariah et~al.(2011)Visweswariah, Rajkumar, Gandhe, Ramanathan,
  and Navratil}]{visweswariah-etal-2011-word}
Karthik Visweswariah, Rajakrishnan Rajkumar, Ankur Gandhe, Ananthakrishnan
  Ramanathan, and Jiri Navratil. 2011.
\newblock \href {https://www.aclweb.org/anthology/D11-1045} {{A Word Reordering
  Model for Improved Machine Translation}}.
\newblock In \emph{Proceedings of the 2011 Conference on Empirical Methods in
  Natural Language Processing}, pages 486--496, Edinburgh, Scotland, UK.
  Association for Computational Linguistics.

\end{thebibliography}
\bibliographystyle{acl_natbib}

\appendix

\clearpage

\section{Hyperparameters and implementation of the LMs}
\label{sec:appendix_hyper}

\begin{table*}[t]
    \centering
    {\small
    \begin{tabular}{llc} \toprule
     \multirow{2}{*}{Fairseq model} & architecture & transformer\_lm \\
      & adaptive softmax cut off & 50,000, 140,000 \\
    \cmidrule(lr){1-1} \cmidrule(lr){2-2} \cmidrule(lr){3-3}
    \multirow{5}{*}{Optimizer} & algorithm & Nesterov accelerated gradient (nag) \\
    & learning rates & 1e-5 \\
    & momentum & 0.99 \\
    & weight decay & 0 \\
    & clip norm & 0.1 \\
    \cmidrule(lr){1-1} \cmidrule(lr){2-2} \cmidrule(lr){3-3}
    \multirow{9}{*}{Learning rate scheduler} & type & cosine \\
    & warmup updates & 16,000 \\
    & warmup init lrarning rate & 1e-7 \\
    & max learning rate & 0.1 \\
    & min learning rate & 1e-9 \\
    & t mult (factor to grow the length of each period) & 2 \\
    & learning rate period updates & 270,000 \\
    & learning rate shrink & 0.75 \\
    \cmidrule(lr){1-1} \cmidrule(lr){2-2} \cmidrule(lr){3-3}
    \multirow{1}{*}{Training} & batch size & 4608 tokens \\ 
    & epochs & 3 \\  \bottomrule
        \end{tabular}
        }
        \caption{Hyperparameters of the LMs.}
        \label{tbl:hyperparam}
\end{table*}

We used the Transformer~\cite{vaswani2017} LMs implemented in fairseq~\cite{ott2019fairseq}.
Table~\ref{tbl:hyperparam} shows the hyperparameters of the LMs.
The adaptive softmax cutoff \cite{armand-adaptive-softmax} is only applied to SLM.
We split 10K sentences for dev set.
The left-to-right and right-to-left CLMs achieved a perplexity of 11.05 and 11.08, respectively.
The left-to-right and right-to-left SLMs achieved a perplexity of 28.51 and 28.25, respectively.
Note that the difference in the perplexities between CLM and SLM is due to the difference in the vocabulary size.

\section{Details on Section~\ref{subsec:cons_double} (double objects)}
\label{sec:appendix_detail_exp2}

\noindent
\subsection{Word order for each verb}
\label{subsec:appendix_word_order_and_each_verb}
It is considered that different verbs have different preferences in the order of their object.
For example, while the verb ``{\it 例える}" ({\it compare}) prefers the \texttt{ACC-DAT} order (Example (9)-a), the verb ``{\it 表する}" ({\it express}) prefers the \texttt{DAT-ACC} order (Example (9)-b).

{\small
\eenumsentence[9]{
    \item \shortexnt{3}{{\it 人間を} & {\it 色に} & {\it {\bf 例えた}}.} {person-\texttt{ACC} & color-\texttt{DAT} & compared.} \\ {($\phi_{I}$ compared a person to color.)}
    \item \shortexnt{3}{{\it 店主に}  & {\it 敬意を} & {\it {\bf 表した}}.} {shopkeeper-\texttt{DAT} & respect-\texttt{ACC} & expressed.} \\ {($\phi_{I}$ expressed a respect to a shopkeeper.)}
    }
}

\noindent
Table~\ref{tbl:top5} shows the verbs with the top five and the five worst $R_{\texttt{ACC-DAT}}^v$.

\begin{table*}[t]
    \centering
    {\small   
    \begin{tabular}{llcclcc} \toprule
    & \multicolumn{3}{c}{\texttt{ACC-DAT} is preferred} & \multicolumn{3}{c}{\texttt{DAT-ACC} is preferred} \\
    Model & Verb & $R_{\texttt{ACC-DAT}}^v$ & S\&O & Verb & $R_{\texttt{ACC-DAT}}^v$ & S\&O \\
            \cmidrule(lr){1-1} \cmidrule(lr){2-4} \cmidrule(lr){5-7}
    \multirow{5}{*}{CLM} & ``例える'' ({\it compare}) & 0.993 & 0.945 & ``表する'' ({\it to table}) & 0.001 & 0.013 \\
& ``換算する'' ({\it converted}) & 0.992 & 0.935 & ``澄ます'' ({\it put on airs}) & 0.000 & 0.017 \\
& ``押し出す'' ({\it extruded}) & 0.979 & 0.923 & ``煮やす'' ({\it cook inside}) & 0.000 & 0.019 \\
& ``見立てる'' ({\it mitateru}) & 0.994 & 0.919 & ``瞑る'' ({\it close the eyes}) & 0.001 & 0.021 \\
& ``変換'' ({\it conversion}) & 0.975 & 0.898 & ``竦める'' ({\it shrug}) & 0.002 & 0.022 \\
            \cmidrule(lr){1-1} \cmidrule(lr){2-4} \cmidrule(lr){5-7}
    \multirow{5}{*}{SLM}& ``例える'' ({\it compare}) & 0.993 & 0.926 & ``喫する'' ({\it kissuru}) & 0.003 & 0.018 \\
& ``押し出す'' ({\it extruded}) & 0.979 & 0.914 & ``表する'' ({\it to table}) & 0.001 & 0.018 \\
& ``監禁'' ({\it confinement}) & 0.885 & 0.912 & ``澄ます'' ({\it put on airs}) & 0.000 & 0.021 \\
& ``役立てる'' ({\it help}) & 0.933 & 0.904 & ``抜かす'' ({\it leave out}) & 0.002 & 0.022 \\
& ``帰す'' ({\it attributable}) & 0.838 & 0.903 & ``踏み入れる'' ({\it step into}) & 0.002 & 0.025 \\ \bottomrule
        \end{tabular}
        }
        \caption{The verbs with the top five and the worst five $R_{\texttt{ACC-DAT}}^v$ in each LM. The ``S\&O" columns show the \texttt{ACC-DAT} rate reported in~\citet{sasano2016corpus}.}
        \label{tbl:top5}
\end{table*}

\vspace{\baselineskip}
\noindent
\subsection{Word order and verb types}
\label{subsec:appendix_word_order_and_verb_type}
There are two types of causative-inchoative alternating verbs in Japanese: show-type verbs and pass-type verbs.
The verb types are determined by the subject of the sentence where the corresponding inchoative verb is used.
For the show-type verbs, the \texttt{DAT} argument of a causative sentence becomes the subject in its corresponding inchoative sentence (Example (10)).
On the other hand, the \texttt{ACC} argument of a causative sentence becomes the subject in its corresponding inchoative sentence for the pass-type verbs (Example (11)).

{\small
\begin{description}
\item[\textmd{(10)}] \begin{description}
    \item[Causative: ] \gl{{\it \underline{生徒に}}}{student-\texttt{DAT}} \gl{{\it 本を}}{book-\texttt{ACC}} \gl{{\it {\bf 見せた}}}{showed.} \\
    ($\phi_{I}$ showed a student a book.)
    \vspace{0.15cm}
    \item[Inchoative: ] \gl{{\it \underline{生徒が}}}{student-\texttt{NOM}} \gl{{\it {\bf 見た}}}{saw.} \\ (A student saw $\phi_{\text{something}}$.)
\end{description}
\end{description}
}

{\small
\begin{description}
\item[\textmd{(11)}] \begin{description}
    \item[Causative: ] \gl{{\it \underline{生徒に}}}{student-\texttt{DAT}} \gl{{\it 本を}}{book-\texttt{ACC}} \gl{{\it {\bf 渡した}}}{showed.} \\
    ($\phi_{I}$ passed a student a book.)
    \vspace{0.15cm}
    \item[Inchoative: ] \gl{{\it \underline{本が}}}{book-\texttt{NOM}} \gl{{\it {\bf 渡った}}}{passed.} \\ (A book passed to $\phi_{\text{something}}$.)
\end{description}
\end{description}
}

\noindent
\citet{matsuoka2003two} claims that the show-type verb prefers the \texttt{DAT-ACC} order, while the pass-type verb prefers the \texttt{ACC-DAT} order.

Table~\ref{tbl:show_pass} shows $R_{\texttt{ACC-DAT}}^v$ of the show-type and pass-type verbs.
The results show no significant difference in word order trends between show-type and pass-type verbs, which are consistent with that of~\citet{sasano2016corpus}.

\begingroup
\begin{table*}[t]
\centering
\renewcommand{\arraystretch}{0.5}
    {\scriptsize
    \begin{tabular}{lccclccclccc} \toprule
    \multicolumn{4}{c}{Show-type} & \multicolumn{8}{c}{Pass-type} \\
    \cmidrule(lr){1-4} \cmidrule(lr){5-12}
    Verb & CLM & SLM & S\&O & Verb & CLM & SLM & S\&O & Verb & CLM & SLM & S\&O \\
    \cmidrule(lr){1-4} \cmidrule(lr){5-8} \cmidrule(lr){9-12}
    ``{\it 知らせる}'' ({\it notify}) & .718 & .754 & .522 & ``{\it 戻す}'' ({\it put back}) & .366 & .395 & .771 & ``{\it 漏らす}'' ({\it leak}) & .152 & .207 & .332 \\
    ``{\it 預ける}'' ({\it deposit}) & .426 & .391 & .399 & ``{\it 止める}'' ({\it lodge}) & .638 & .704 & .748 & ``{\it 浮かべる}'' ({\it float}) & .387 & .406 & .255  \\
    ``{\it 見せる}'' ({\it show}) & .353 & .429 &  .301 & ``{\it 包む}'' ({\it wrap}) & .316 & .356 & .603 & ``{\it 向ける}'' ({\it direct}) & .291 & .319 & .251 \\
    ``{\it 被せる}'' ({\it cover}) & .240 & .224 & .256 & ``{\it 伝える}'' ({\it inform}) & .419 & .460 & .522 & ``{\it 残す}'' ({\it leave}) & .323 & .318 & .238 \\
    ``{\it 教える}'' ({\it teach}) & .297 & .293 & .235 & ``{\it 乗せる}'' ({\it place on}) & .556 & .498 &.496 & ``{\it 埋める}'' ({\it bury}) & .405 & .430 & .223 \\
    ``{\it 授ける}'' ({\it give}) & .101 & .084 & .186 & ``{\it 届ける}'' ({\it deliver}) & .364 & .419 & .491 & ``{\it 混ぜる}'' ({\it blend}) & .336 & .276 & .200 \\
    ``{\it 浴びせる}'' ({\it shower}) & .113 & .121 &.177 & ``{\it 並べる}'' ({\it range}) & .423 & .485 &.481 & ``{\it 当てる}'' ({\it hit}) & .287 & .320 & .185 \\
    ``{\it 貸す}'' ({\it lend}) & .253 & .213 & .118 & ``{\it ぶつける}'' ({\it knock}) & .333 & .344 & .436 & ``{\it 掛ける}'' ({\it hang}) & .285 & .288 & .108 \\
    ``{\it 着せる}'' ({\it dress}) & .115 & .109 & .113 & ``{\it 付ける}'' ({\it attach}) & .326 & .329 & .368 & ``{\it 重ねる}'' ({\it pile}) & .226 & .263 & .084 \\
    - & - & - & - & ``{\it 渡す}'' ({\it pass}) & .349 & .336 & .362 & ``{\it 建てる}'' ({\it build}) & .117 & .099 & .069 \\ 
    - & - & - & - & ``{\it 落とす}'' ({\it drop}) & .379 & .397 & .351 & - & - & - & - \\
    \cmidrule(lr){1-4} \cmidrule(lr){5-12}
    Macro Avg. & .291 & .291 & .305 & \multicolumn{5}{c}{Macro Avg.} & .347 & .364 & .361 \\ \bottomrule
    \end{tabular}
    }
    \caption{Overlap of the results of LMs and that of~\citet{sasano2016corpus} on the relationship of the \texttt{ACC-DAT} rate and verb types. Each score corresponding to a verb denotes its \texttt{DAT-ACC} rate. The ``S\&O" columns show the \texttt{ACC-DAT} rate reported in~\citet{sasano2016corpus}. There is no significant difference between the distributions of the \texttt{DAT-ACC} rate in two verb types.}
    \label{tbl:show_pass}
\end{table*}
\endgroup

\vspace{\baselineskip}
\noindent
\subsection{Word order and semantic role of the dative argument}
\label{subsec:appendix_word_order_and_semantic_role_dative_argument}
As described in Section~\ref{subsec:cons_double}, \citet{sasano2016corpus} reported that type-A examples prefer the  \texttt{ACC-DAT} order and type-B examples prefer the \texttt{DAT-ACC} order.
We used the same examples as~\citet{sasano2016corpus} used.
We analyzed the difference in the trend of argument order between type-A and type-B examples in each verb.
Table~\ref{tbl:animacy} shows the verbs, which show a significant change in the argument order between type-A and type-B examples (p $<$ 0.05 in a two-proportion z-test).
In the experiment using CLM, 31 verbs show the trend that type-A examples more prefer the \texttt{ACC-DAT} order to type-B, and 17 verbs show contrary trends.
In the experiment using SLM, 38 verbs show the trend that type-A examples more prefer the \texttt{ACC-DAT} order to type-B, and 11 verbs show contrary trends.
These results show that the number of verbs, where the \texttt{ACC-DAT} order is preferred by type-A examples rather than type-B, is significantly larger (p $<$ 0.05 with a two-sided sign test).
This experimental design follows~\citet{sasano2016corpus}.

\begin{table*}[t]
    {\small
    \begin{tabular}{lp{7cm}p{7cm}} \toprule
    Model & Verbs whose type-A examples prefer the \texttt{ACC-DAT} order & Verbs whose type-B examples prefer the  \texttt{ACC-DAT} order \\
        \cmidrule(lr){1-1} \cmidrule(lr){2-2} \cmidrule(lr){3-3}
    CLM & ``預ける'' ({\it deposit}), ``置く'' ({\it put}), ``持つ'' ({\it to have}), ``入れる'' ({\it put in}), ``納める'' ({\it pay}), ``郵送'' ({\it mailing}), ``供給'' ({\it supply}), ``出す'' ({\it put out}), ``運ぶ'' ({\it transport}), ``流す'' ({\it shed}), ``掛ける'' ({\it multiply}), ``飾る'' ({\it decorate}), ``広げる'' ({\it spread}), ``移す'' ({\it transfer}), ``残す'' ({\it leave}), ``配送'' ({\it delivery}), ``送る'' ({\it send}), ``投げる'' ({\it throw}), ``送付'' ({\it sending}), ``返却'' ({\it return}), ``届ける'' ({\it deliver}), ``戻す'' ({\it return}), ``着ける'' ({\it wear}), ``上げる'' ({\it increase}), ``落とす'' ({\it drop}), ``載せる'' ({\it load}), ``変更'' ({\it change}), ``納入'' ({\it delivery}), ``卸す'' ({\it sell ​​wholesale}), ``掲載'' ({\it published}), ``通す'' ({\it through}) & ``配布'' ({\it distribution}), ``渡す'' ({\it hand over}), ``プレゼント'' ({\it present}), ``合わせる'' ({\it match}), ``見せる'' ({\it show}), ``提供'' ({\it offer}), ``与える'' ({\it give}), ``当てる'' ({\it hit}), ``回す'' ({\it turn}), ``追加'' ({\it add to}), ``貸す'' ({\it lend}), ``展示'' ({\it exhibition}), ``据える'' ({\it lay}), ``依頼'' ({\it request}), ``挿入'' ({\it insertion}), ``纏める'' ({\it collect}), ``請求'' ({\it claim}) \\
        \cmidrule(lr){1-1} \cmidrule(lr){2-2} \cmidrule(lr){3-3}
    SLM & ``預ける'' ({\it deposit}), ``置く'' ({\it put}), ``頼む'' ({\it ask}), ``入れる'' ({\it put in}), ``納める'' ({\it pay}), ``郵送'' ({\it mailing}), ``出す'' ({\it put out}), ``運ぶ'' ({\it transport}), ``流す'' ({\it shed}), ``掛ける'' ({\it multiply}), ``広げる'' ({\it spread}), ``移す'' ({\it transfer}), ``残す'' ({\it leave}), ``リクエスト'' ({\it request}), ``配送'' ({\it delivery}), ``送る'' ({\it send}), ``投げる'' ({\it throw}), ``送付'' ({\it sending}), ``求める'' ({\it ask}), ``提出'' ({\it submission}), ``届ける'' ({\it deliver}), ``要求'' ({\it request}), ``戻す'' ({\it return}), ``寄付'' ({\it donation}), ``寄贈'' ({\it donation}), ``着ける'' ({\it wear}), ``乗せる'' ({\it place}), ``上げる'' ({\it increase}), ``落とす'' ({\it drop}), ``貼る'' ({\it stick}), ``分ける'' ({\it divide}), ``ばらまく'' ({\it spamming}), ``はめる'' ({\it fit}), ``支払う'' ({\it pay}), ``配達'' ({\it delivery}), ``卸す'' ({\it sell ​​wholesale}), ``纏める'' ({\it collect}), ``通す'' ({\it through}) & ``プレゼント'' ({\it present}), ``持つ'' ({\it to have}), ``合わせる'' ({\it match}), ``見せる'' ({\it show}), ``向ける'' ({\it point}), ``提供'' ({\it offer}), ``装備'' ({\it equipment}), ``追加'' ({\it add to}), ``展示'' ({\it exhibition}), ``据える'' ({\it lay}), ``採用'' ({\it adopt}) \\ \bottomrule
        \end{tabular}
        }
        \caption{The verbs which show a significant change in the argument order trend depending on the semantic role of its dative argument. The scores denote the \texttt{DAT-ACC} rate. Type-A corresponds to the examples with an inanimate goal dative argument. Type-B corresponds to the examples with an animate processor dative argument. The number of type-A verbs is significantly larger than that of type-B verbs.}
        \label{tbl:animacy}
\end{table*}

\vspace{\baselineskip}
\noindent
\subsection{Word order and co-occurrence of verb and arguments}
\label{subsec:appendix_word_order_and_co-occurrence_of_verb_and_arguments}
We evaluate the claim that an argument frequently co-occurring with the verb tends to be placed near the verb.
We examine the relationship between each example's word order trend and $\Delta$NPMI.
$\Delta$NPMI is calculated as follows:

\vspace{-0.3cm}
\begin{align*}
\Delta \text{NPMI} &= \text{NPMI}(n_{\texttt{DAT}}, v) \\ &\:\:\:\:-  \text{NPMI}(n_{\texttt{ACC}}, v) \:\:, \\
\text{where} \: \text{NPMI}(n_c, v) &= \frac{\text{PMI}(n_c, v)}{-log(p(n_c, v))} \:\:, \\
\text{PMI}(n_c, v) &= log\frac{p(n_c, v)}{p(n_c)p(v)} \:\:,
\end{align*}

\normalsize
\noindent
where, $v$ is a verb and $n_c$ ($c \in$ {\texttt{DAT}, \texttt{ACC}}) is its argument.

\section{Data used in Section~\ref{subsec:place_time}, Section~\ref{sec:topic}, and Appendix~\ref{sec:appendix_add_ana}}
\label{sec:appendix_Wikipedia_data}

\begin{table}[t]
    \centering
    {\small   
    \begin{tabular}{lr} \toprule
         Case & \#occurrence \\ 
        \cmidrule(lr){1-1} \cmidrule(lr){2-2} 
        \texttt{TIM} & 11,780 \\
        \texttt{LOC} & 15,544 \\
        \texttt{NOM} & 55,230 \\
        \texttt{DAT} & 56,243 \\ 
        \texttt{ACC} & 57,823 \\ 
        \bottomrule
        \end{tabular}
        }
        \caption{The number of occurrence for each case in the data used in Section~\ref{subsec:place_time}, Section~\ref{sec:topic}, and Appendix~\ref{sec:appendix_add_ana}}
        \label{tbl:wiki_cases}
\end{table}

First, we randomly collected 50M sentences from 3B web pages.
Note that there is no overlap between the collected sentences and the training data of LMs.
Next, we obtained the sentences that satisfy the following criteria:

\vspace {-0.2 cm}
\begin {itemize}
\setlength { \parskip} {0 cm}
\setlength { \itemsep} {0 cm}
\item There is a verb (placed at the end of the sentence) with more than two arguments (accompanying the case particle {\it ga}, {\it o}, {\it ni}, or {\it de}), where dependency distance between the verb and arguments is one.
\item Each argument (with its descendant) has fewer than 11 morphemes in the argument.
\end {itemize}

\noindent
In each example, the verb (satisfying the above condition), its arguments, and the descendants of the arguments are extracted.
Example sentences are created by concatenating the verb, its argument, and the descendants of the arguments with preserving their order in the original sentences.

In the experiments in Section~\ref{subsec:place_time}, we analyzed the word order trend of the \texttt{TIM} and \texttt{LOC} constituents.
We regard the constituent (argument and its descendants) satisfying the following condition as the \texttt{TIM} constituent:

\vspace {-0.2 cm}
\begin {itemize}
\setlength { \parskip} {0 cm}
\setlength { \itemsep} {0 cm}
\item Accompanying the postpositional case particle ``{\it に}" (\texttt{DAT}).
\item Containing time category morphemes\footnote{\label{note:id_JUMAN}identified by JUMAN}.
\end {itemize}

\noindent
We regard the constituent (argument and its descendants) satisfying the following condition as the \texttt{LOC} constituent:

\vspace {-0.2 cm}
\begin {itemize}
\setlength { \parskip} {0 cm}
\setlength { \itemsep} {0 cm}
\item Accompanying the postpositional case particle ``{\it で}".
\item Containing location category morphemes\footref{note:id_JUMAN}.
\end {itemize}

81k examples were created.
The averaged number of characters in a sentence was 45.1 characters.
The number of occurrences of each case is shown in Table~\ref{tbl:wiki_cases}.
The scrambling process conducted in the experiments (Sections~\ref{subsec:place_time} and \ref{sec:topic}) is the same as described in Section~\ref{sec:correl}.

\section{Details on Section~\ref{subsec:adverb} (adverb)}

\begin{table*}[t]
    \centering
    \renewcommand{\arraystretch}{0.8}
    {\small   
    \begin{tabular}{lcccccccc} \toprule
         \multirow{2}{*}{Model} & \multicolumn{2}{c}{\textsc{Modal}} & \multicolumn{2}{c}{\textsc{Time}} & \multicolumn{2}{c}{\textsc{Manner}} & \multicolumn{2}{c}{\textsc{Resultive}} \\ 
         & Canonical & $r$ & Canonical & $r$ & Canonical & $r$ & Canonical & $r$ \\
        \cmidrule(lr){1-1} \cmidrule(lr){2-3} \cmidrule(lr){4-5} \cmidrule(lr){6-7} \cmidrule(lr){8-9}
        CLM  & ASOV & 1. & ASOV, SAOV & 1. & SAOV, SOAV & 0.5 & SAOV, SOAV & 1. \\
        SLM & ASOV & 1. & SAOV & 0.5 & SAOV, SOAV & 1. & SOAV  & 0.5 \\
        Koizumi(2016) & ASOV & - & ASOV, SAOV & - & SAOV, SOAV & - & SAOV, SOAV & - \\ \bottomrule
        \end{tabular}
        }
        \caption{Overlap of the preference of the adverb position of LMs and that of~\citet{koizumi2006}. The column ``Canonical'' shows the adverb position, which is significantly preferred over the other positions. The score $r$ denotes the Pearson correlation coefficient of the preferred rank of three possible adverb positions obtained from LMs to that of~\citet{koizumi2006}. }
        \label{tbl:adverb_detail}
\end{table*}

Table~\ref{tbl:adverb_detail} shows the correlation between the result of LMs and that of~\citet{koizumi2006}.
The column ``Canonical'' shows the position, which is significantly preferred over the other positions. ``A,'' ``S,'' ``O,'' and ``V'' denote ``adverb,'' ``subject,'' ``object,'' and ``verb,'' respectively. 
The sequence of the alphabets corresponds to their order; for example, ``ASOV'' indicates the order: adverb $<$ subject $<$ object $<$ verb. 
Following~\citet{koizumi2006}, we examined the three candidate positions of the adverb: ``ASOV," ``SAOV," and ``SOAV."
The score $r$ denotes the Pearson correlation coefficient of the preferred ranks of each adverb position to that reported in~\citet{koizumi2006}.

\section{Details on Section~\ref{subsec:topic:exp} (topicalization)}
\label{sec:appendix_detail_ana}

\begin{table}[t]
    {\small
    \begin{center}
    \subfloat[][CLM]{
    \begin{tabular}{lccccc} \toprule
&\texttt{TIM}&	\texttt{PLC}&	\texttt{NOM}&	\texttt{DAT}&	\texttt{NOM} \\
\cmidrule(lr){1-1} \cmidrule(lr){2-6}
\texttt{TIM} &	-&	.490& 	.329& 	.720& 	.698 \\ 
\texttt{PLC} &	.510& 	-&	.484& 	.748& 	.742 \\ 
\texttt{NOM} &	.671& 	.516& 	-&	.804& 	.852 \\ 
\texttt{DAT} &	.280& 	.252& 	.196& 	-&	.536 \\ 
\texttt{NOM} &	.302& 	.258& 	.148& 	.464& 	- \\ \bottomrule
    \end{tabular}
    }
    \hspace{1cm}
    \subfloat[][SLM]{
    \begin{tabular}{lccccc} \toprule
&\texttt{TIM}&	\texttt{PLC}&	\texttt{NOM}&	\texttt{DAT}&	\texttt{NOM} \\
\cmidrule(lr){1-1} \cmidrule(lr){2-6}
\texttt{TIM} &	-&	.538& 	.402& 	.676& 	.711 \\ 
\texttt{PLC} &	.462 & 	-&	.553 & 	.757 & 	.749 \\ 
\texttt{NOM} &	.598 & 	.447 & 	-&	.774& 	.834 \\ 
\texttt{DAT} &	.324& 	.243& 	.226& 	-&	.552 \\ 
\texttt{NOM} &	.289& 	.251& 	.166& 	.448& 	- \\ \bottomrule
    \end{tabular}
     }
    \end{center}
    }
    \caption{The scores denote $t_{a|b}$. The row corresponds to the case $a$, the column corresponds to $b$. Higher $t_{a|b}$ suggests the trend that the case $a$ is more likely to be topicalized than the case $b$.}
    \label{tbl:topicalization_case}
\end{table}

\begin{table}[t]
    \centering
    {\small   
    \begin{tabular}{lp{3cm}} \toprule
         Original case particle & After the adverbial particle ``{\it は}" (\texttt{TOP}) is added \\ 
        \cmidrule(lr){1-1} \cmidrule(lr){2-2} 
        {\it が} (\texttt{TOP}) & {\it \dout{が}は} \\
        {\it に} (\texttt{TIM}, \texttt{DAT}) & {\it には} \\
        {\it を} (\texttt{ACC}) & {\it \dout{を}は} \\
        {\it で} (\texttt{LOC}) & {\it では} \\ \bottomrule
        \end{tabular}
        }
        \caption{Rules of deleting the original case particle when the adverbial particle ``{\it は}'' (\texttt{TOP}) is added. This rule is also applied when adding the other adverbial particles (Appendix~\ref{sec:appendix_add_ana}).}
        \label{tbl:topicalize_rule}
\end{table}

We topicalized a specific constituent by moving the constituent to the beginning of the sentence and adding the adverbial particle ``{\it は}" (\texttt{TOP}).
Strictly speaking, conjunctions are preferentially placed at the beginning of the sentence rather than topicalized constituents.
The examples we used do not include the conjunctions at the beginning of the sentence.
The adverbial particle was added according to the rules shown in Table~\ref{tbl:topicalize_rule}.

\noindent
{\bf Claim (i):}
Table~\ref{tbl:topicalization_case} shows the $t_{a|b}$ for each pair of the case $a$ (row) and $b$ (column).
The results show that the more anterior the case $a$ is and the more posterior the case $b$ is in the canonical word order, the larger the $t_{a|b}$ is.

\noindent
{\bf Claim (ii):}
Figure~\ref{fig:topicalization_verb} shows that the more a verb prefers the \texttt{ACC-DAT} order, the more \texttt{ACC} case tends to be topicalized.
The X-axis denotes the \texttt{ACC-DAT} rate of the verb, and the Y-axis denotes the trend that \texttt{ACC} is more likely to be topicalized than \texttt{DAT}.

{\setlength\textfloatsep{0pt}
\begin{figure}[t]
    \centering
      \includegraphics[width=6.5cm]{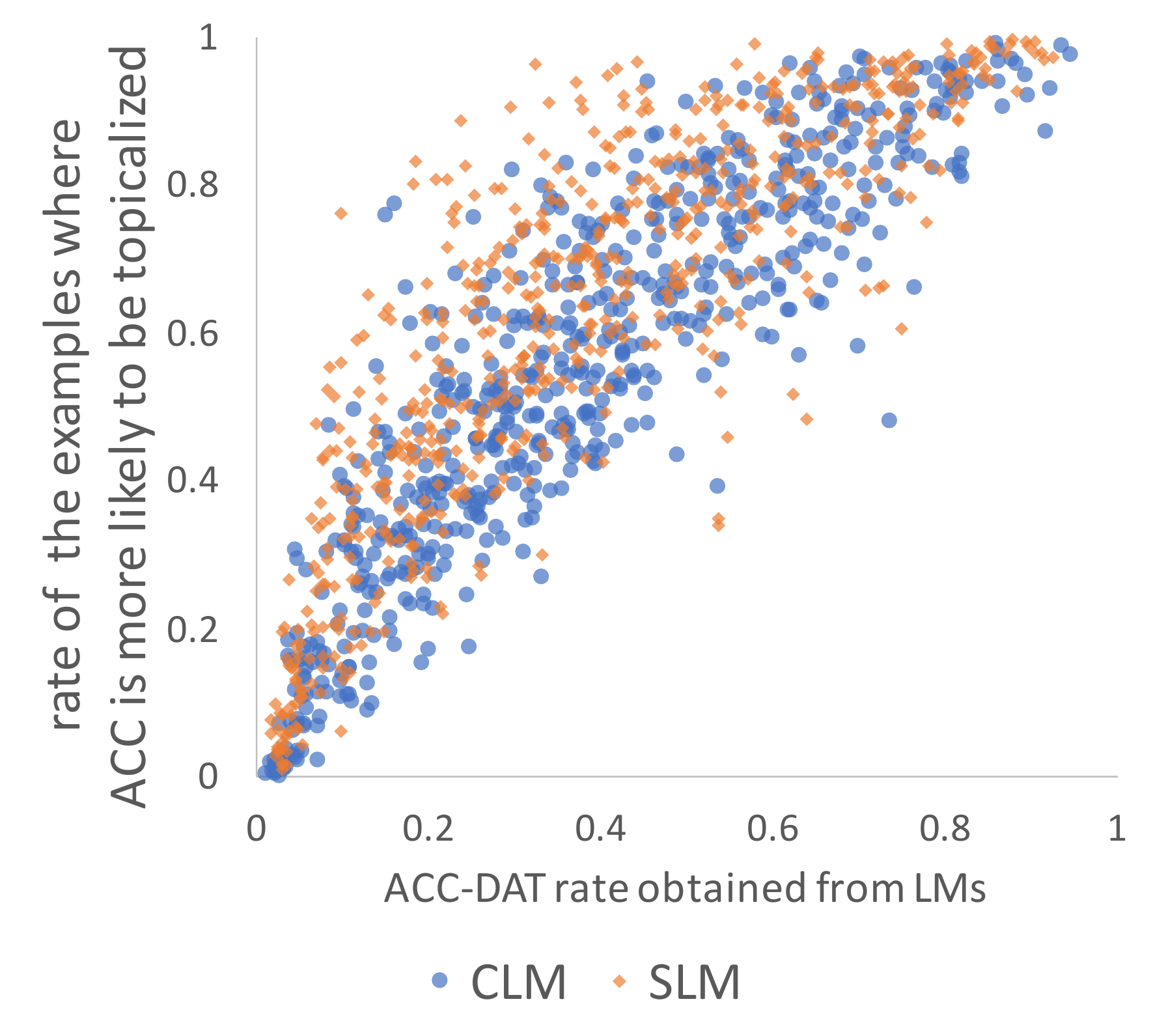}
      \vspace{-0.1cm}
      \caption{Correlation between the \texttt{ACC-DAT} rate and the rate that the \texttt{ACC} argument is more likely to be topicalized than \texttt{DAT} for each verb. Each plot corresponds to the result of each verb.}
      \label{fig:topicalization_verb}
\end{figure}
}

\section{Additional analysis: adverbial particles and their effect for word order}
\label{sec:appendix_add_ana}

\paragraph*{The adverbial particles}
We can add supplementary information with adverbial particles.
The adverbial particle  ``{\it は}" (\texttt{TOP}) is the typical one.
In Example (12), the adverbial particle ``{\it も}" ({\it also}), instead of ``{\it を}" (\texttt{ACC}), implies that there is another thing the teacher gave to the student ({\it ``a teacher gave not only $\phi$ but also a book to a student."}).

{\small
\enumsentence[(12)]{
    \shortexnt{3}{{\it \uwave{生徒に}}  & {\it \underline{本\dout{を} も}} & {\it {\bf あげた}}.} {student-\texttt{DAT} & also book-\texttt{ACC} & gave.} \\ {}
}
}

\paragraph*{Experiments}
A constituent accompanying the adverbial particle ``{\itは}" ({\texttt{TOP}}) is moved to the beginning of the sentence~\cite{noda1996}.
However, it is not clear whether other adverbial particles also have the above property.
In this section, we evaluate the following claim: a different adverbial particle shows different degrees of the effects for the word order.

For each example $s \in S$ collected from Japanese Wikipedia, we replaced the postpositional particle with a specific adverbial particle, following the rules in Table~\ref{tbl:topicalize_rule}.
We used four typical adverbial particles: ``{\it は}" (\texttt{TOP}), ``{\it こそ}" (emphasis), ``{\it も}" ({\it also}), and ``{\it だけ}" ({\it only}).
Two variants of word order, {\it Non-moved}, and {\it Moved} were created for each example.
Example (13) is an example focusing on the \texttt{ACC} case with the particle ``{\it も}" ({\it also}).

{\scriptsize
\begin{description}
\setlength{\itemindent}{-16pt}
\item[\textmd{(13)}] \begin{description}
    \item[Original: ] \gl{{\it \uwave{生徒に}}}{student-\texttt{DAT}} \gl{{\it \underline{本を}}}{book-\texttt{ACC}} \gl{{\it {\bf あげた}}.}{gave.} \\
    \item[Non-moved: ] \gl{{\it \uwave{生徒に}}}{student-\texttt{DAT}} \gl{{\it \underline{本\dout{を}も}}}{also book-\texttt{ACC}} \gl{{\it {\bf あげた}}.}{gave.} \\
    \item[Moved: ] \gl{{\it \underline{本\dout{を}も}}}{also book-\texttt{ACC}} \gl{{\it \uwave{生徒に}}}{student-\texttt{DAT}} \gl{{\it \underline{\dout{本を}}}}{\dout{book-\texttt{ACC}}} \gl{{\it {\bf あげた}}.}{gave.} \\
\end{description}
\end{description}
}

\begin{table*}[t]
    \centering
    {\small
    \begin{tabular}{lccccccc} \toprule
        Model & Toritate particle & \texttt{TIM} & \texttt{LOC}  & \texttt{NOM} & \texttt{DAT} & \texttt{ACC} & Avg. \\ 
        \cmidrule(lr){1-1} \cmidrule(lr){2-2} \cmidrule(lr){3-7} \cmidrule(lr){8-8} 
        \multirow{4}{*}{CLM} & ``{\it は}" (\texttt{TOP}) & .715 & .777 & .675 & .624 & .623 & .683 \\
        & {``\it こそ}" (emphasis)& .492 & .423 & .521 & .313 & .486 & .447 \\
        & ``{\it も}" ({\it also}) & .560 & .557 & .458 & .343 & .271 & .438 \\
        & ``{\it だけ}" ({\it only}) & .385 & .340 & .312 & .227 & .184 & .331 \\ 
        \cmidrule(lr){2-2} \cmidrule(lr){3-7} \cmidrule(lr){8-8} 
        & Avg. & .538 & .525 & .544 & .377 & .391 & - \\
        \cmidrule(lr){1-1} \cmidrule(lr){2-8}
        \multirow{4}{*}{SLM} & ``{\it は}" (\texttt{TOP}) & .667 & .751 & .635 & .565 & .580 & .640 \\
        & ``{\it こそ}" (emphasis) & .567 & .596 & .574 & .398 & .462 & .519 \\
        & ``{\it も}" ({\it also}) & .511 & .531 & .457 & .292 & .259 & .410 \\
        & ``{\it だけ}" ({\it only}) & .334 & .309 & .285 & .172 & .126 & .303 \\ 
        \cmidrule(lr){2-2} \cmidrule(lr){3-7} \cmidrule(lr){8-8} 
        & Avg. & .520 & .547 & .560 & .357 & .357 & - \\\bottomrule
        \end{tabular}
        }
        \caption{The scores denote that the {\it Moved} order is preferred over the {\it Non-moved} order when the corresponding case (column) accompanies the corresponding particle (row). The trend is different depending on the case and particle.}
        \label{tbl:topic_ease_case}
\end{table*}

\vspace{-0.2cm}
\noindent
We compared the generation probabilities between the {\it Non-moved} and {\it Moved} orders.
We calculated the rate that the {\it Moved} order is preferred in each combination of the case types and the adverbial particles.

\paragraph*{Results}
The results are shown in Table~\ref{tbl:topic_ease_case}.
When using ``{\it は}" (\texttt{TOP}) as a postpositional particle, the {\it Moved} order is preferred to {\it Non-moved}, which is consistent with the well-known characteristics of topicalization described in Section~\ref{sec:topic}.
In addition, the degree of preference between {\it Moved} and {\it Non-moved} differs depending on the adverbial particles. 
Furthermore, the results indicate that the anterior case in the canonical word order is likely to move to the beginning of the sentence by the effect of the adverbial particle.

\paragraph*{Additional experiments and results}
We analyzed the trend of double object order when a specific case accompanies an adverbial particle.
Figure~\ref{fig:topicalization_adv_p_clm} shows the result when the \texttt{ACC} argument accompanies an adverbial particle, and Figure~\ref{fig:topicalization_adv_p_slm} shows the result when the \texttt{DAT} argument accompanies an adverbial particle.
The left parts of these figures show the result of CLM, and the right part of these figures shows the result of SLM.
The X-axis denotes the \texttt{ACC-DAT}$\,$/$\,$\texttt{DAT-ACC} rate of the verb when both of the arguments do not accompany an adverbial particle.
The Y-axis denotes the \texttt{ACC-DAT}$\,$/$\,$\texttt{DAT-ACC} rate when a specific case accompanies an adverbial particle.
The results show that the case accompanying an adverbial particle is likely to be placed near the beginning of the sentence.
In addition, the degree of the above trend depends on the adverbial particles.
These results suggest that some adverbial particles have a effect for word order.

{\setlength\textfloatsep{0pt}
\begin{figure*}[h]
    \centering
    \subfloat[][CLM]{
        \includegraphics[width=7.5cm]{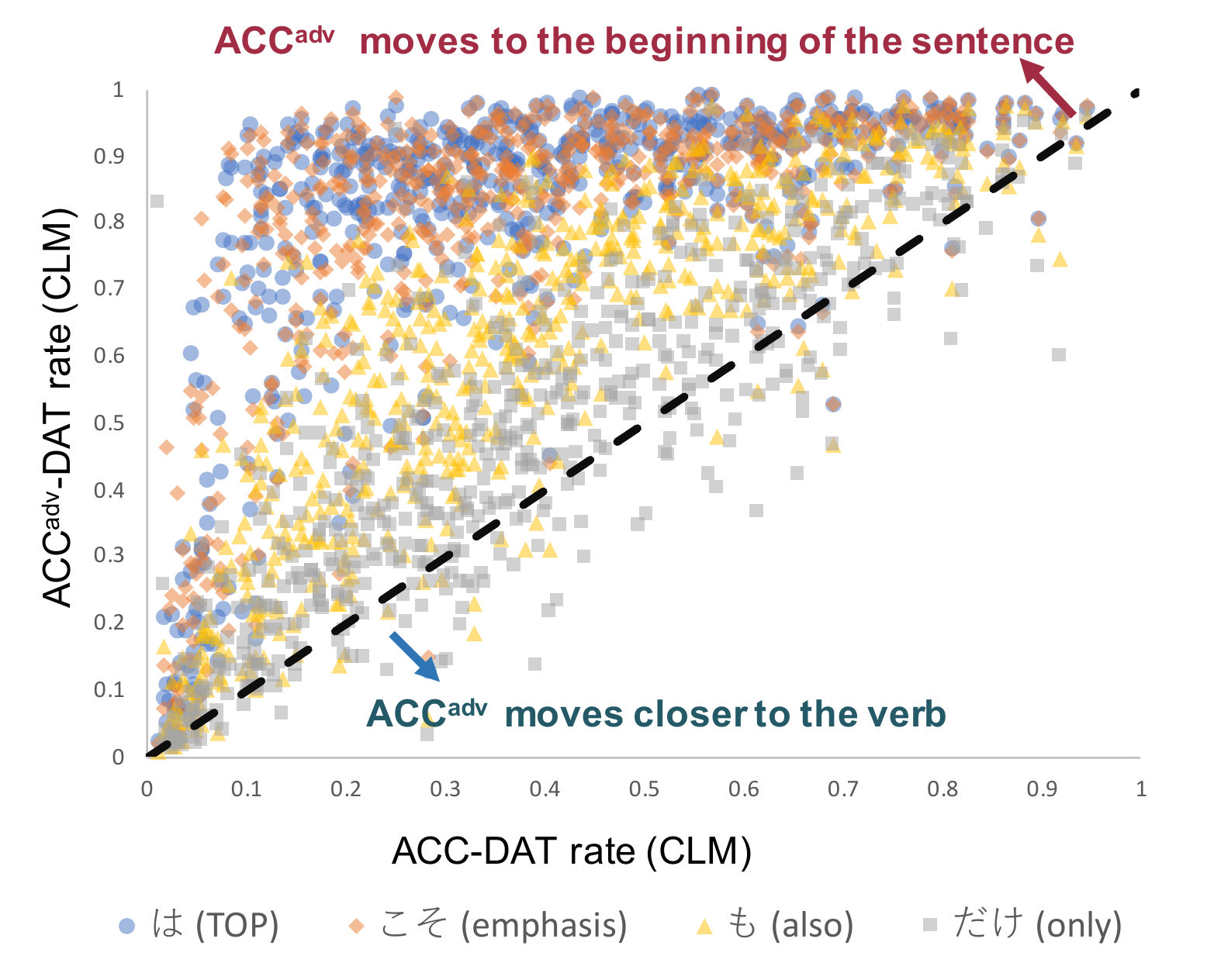}
    }
    \subfloat[][SLM]{
        \includegraphics[width=7.5cm]{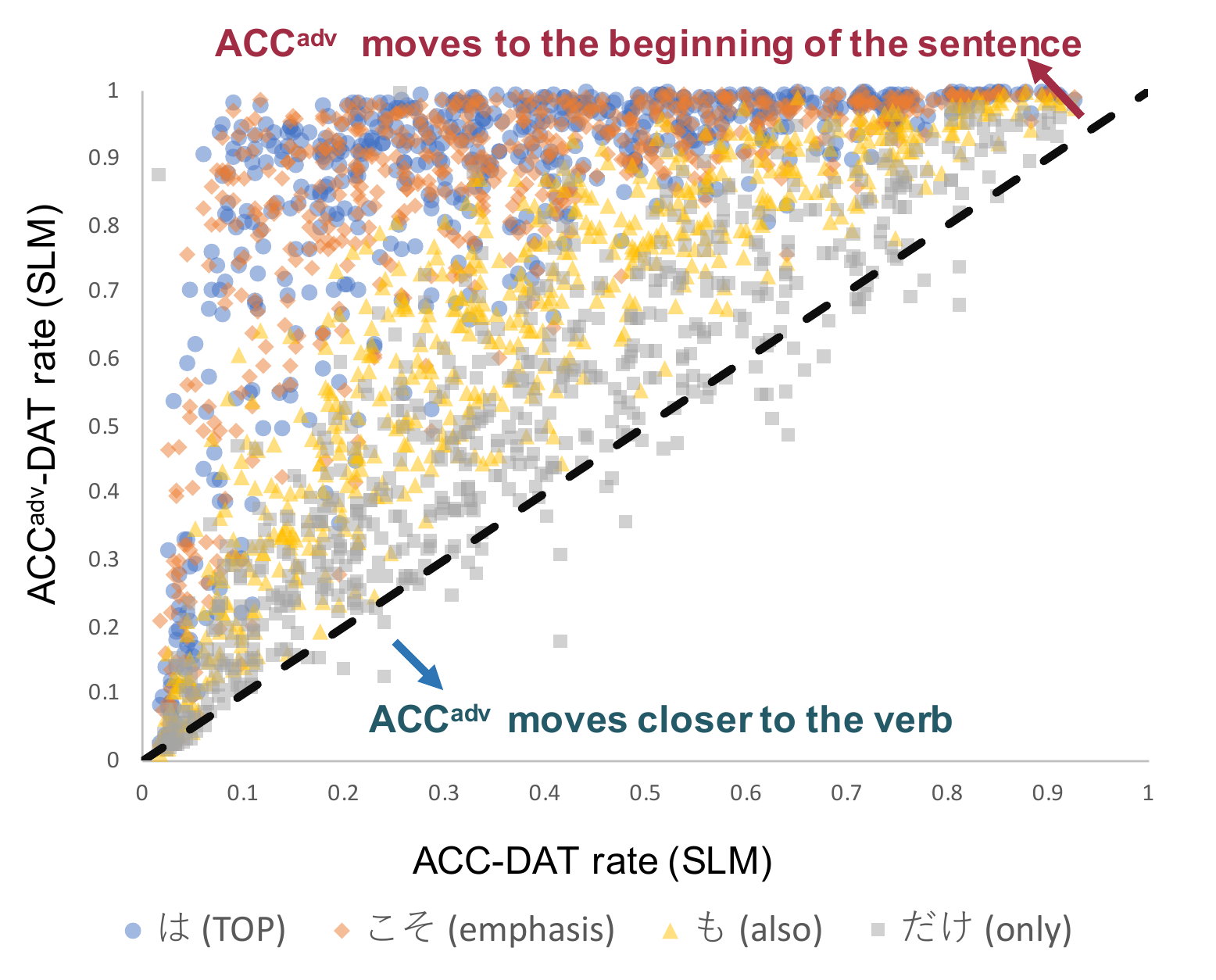}
    }
  \caption{Change of the \texttt{ACC-DAT} order when the \texttt{ACC} argument accompanies an adverbial particle. These results indicate that the \texttt{ACC} argument with an adverbial particle (\texttt{ACC}$^\text{adv}$) is more likely to be placed before the \texttt{DAT} argument. In addition, this trend differs for each particle.}
  \label{fig:topicalization_adv_p_clm}
\end{figure*}
}

{\setlength\textfloatsep{0pt}
\begin{figure*}[h]
    \centering
    \subfloat[][CLM]{
        \includegraphics[width=7.5cm]{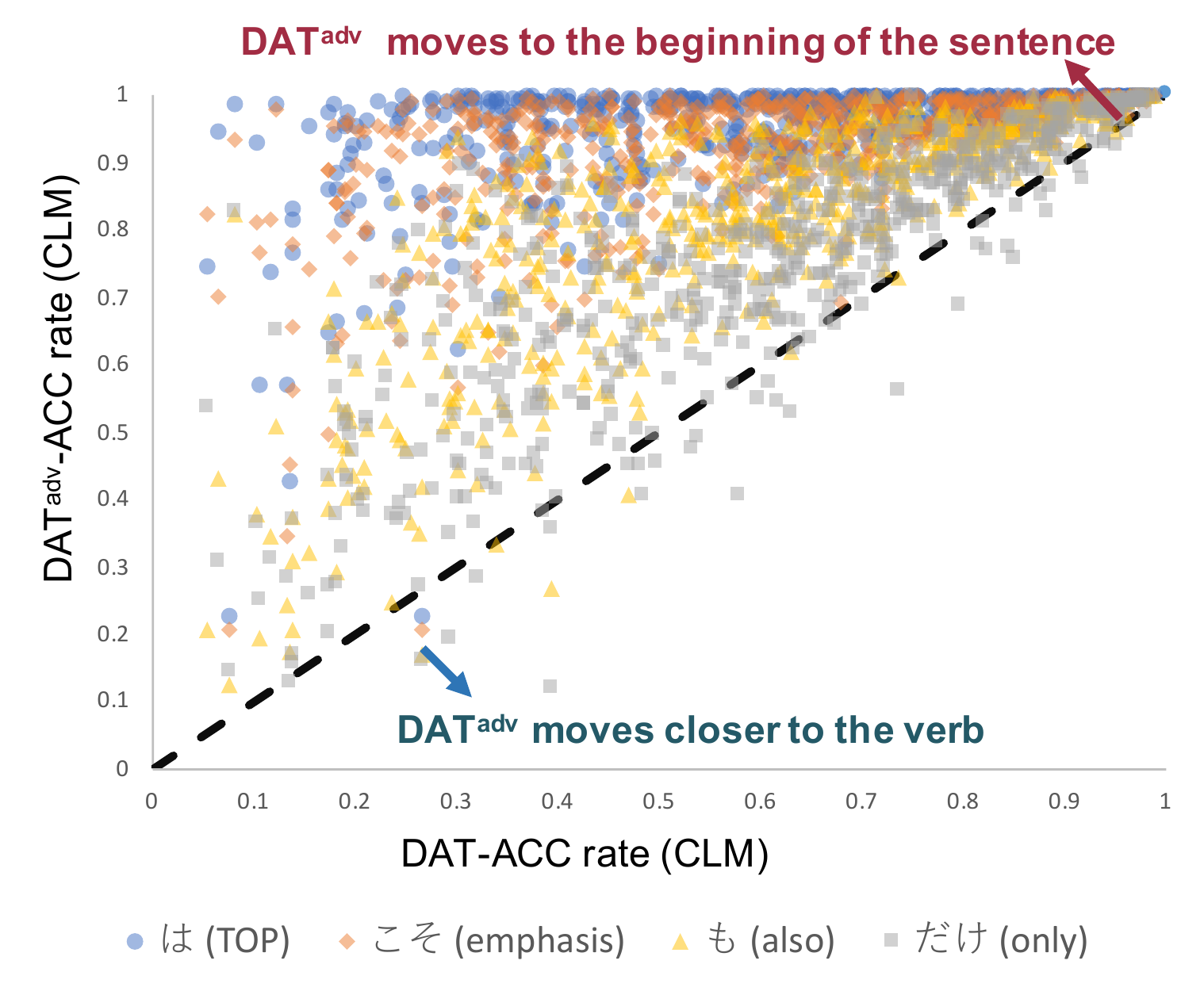}
    }
    \subfloat[][SLM]{
        \includegraphics[width=7.5cm]{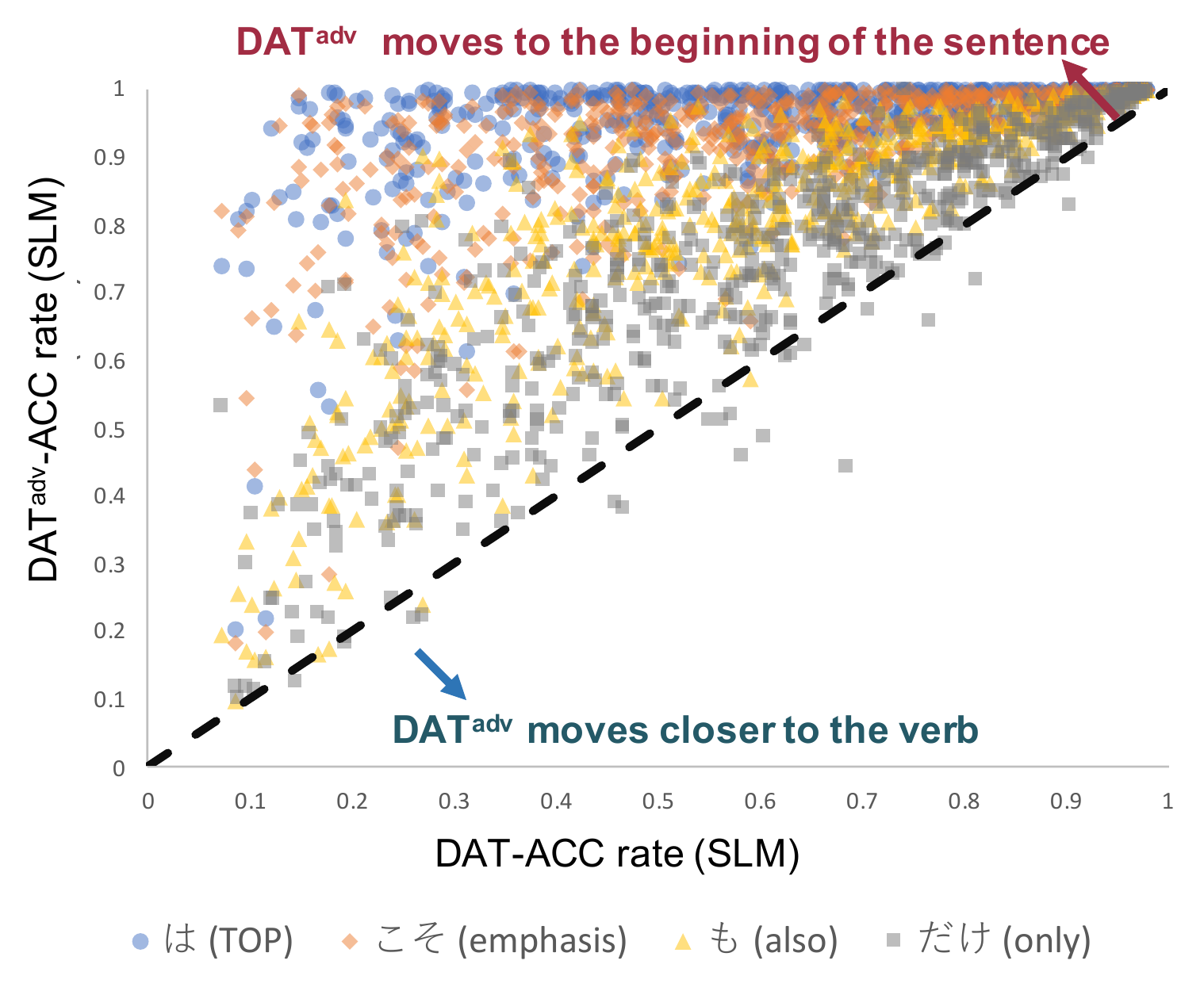}
    }
      \caption{Change of the \texttt{DAT-ACC} order when the \texttt{DAT} argument accompanies an adverbial particle. These results indicate that the \texttt{DAT} argument with an adverbial particle (\texttt{DAT}$^\text{adv}$) is more likely to be placed before the \texttt{ACC} argument. In addition, this trend differs for each particle.}
      \label{fig:topicalization_adv_p_slm}
\end{figure*}
}

\end{document}